%% file: DeepTruthDiscovery_arxiv.tex
\documentclass[10pt,journal,compsoc]{IEEEtran}
\input{preamble.tex}
\usepackage{breqn}
\usepackage{xr}
\usepackage{color}
\usepackage{booktabs}
\usepackage{multirow}

\renewcommand{\hide}[1]{}

\newcommand{\tbC}{\widetilde{\bC}}

\newcommand{\tbmu}{\widetilde{\bmu}}

\newcommand{\tbsigma}{\widetilde{\bm{\sigma}}}

\newcommand{\hbmu}{\widehat{\bm{\mu}}}
\newcommand{\hnabla}{\widehat{\nabla}}
\newcommand{\hbsigma}{\widehat{\bm{\sigma}}}
\newcommand{\onezero}[1]{\mathrm{sum\_one\_hot}\left(#1\right)}

\ifCLASSOPTIONcompsoc
  \usepackage[nocompress]{cite}
\else
  \usepackage{cite}
\fi
\ifCLASSINFOpdf
\else
\fi
\begin{document}
%
% paper title
% Titles are generally capitalized except for words such as a, an, and, as,
% at, but, by, for, in, nor, of, on, or, the, to and up, which are usually
% not capitalized unless they are the first or last word of the title.
% Linebreaks \\ can be used within to get better formatting as desired.
% Do not put math or special symbols in the title.
\title{An Unsupervised Bayesian Neural Network for Truth Discovery in Social Networks}
%
%
% author names and IEEE memberships
% note positions of commas and nonbreaking spaces ( ~ ) LaTeX will not break
% a structure at a ~ so this keeps an author's name from being broken across
% two lines.
% use \thanks{} to gain access to the first footnote area
% a separate \thanks must be used for each paragraph as LaTeX2e's \thanks
% was not built to handle multiple paragraphs
%
%
%\IEEEcompsocitemizethanks is a special \thanks that produces the bulleted
% lists the Computer Society journals use for "first footnote" author
% affiliations. Use \IEEEcompsocthanksitem which works much like \item
% for each affiliation group. When not in compsoc mode,
% \IEEEcompsocitemizethanks becomes like \thanks and
% \IEEEcompsocthanksitem becomes a line break with idention. This
% facilitates dual compilation, although admittedly the differences in the
% desired content of \author between the different types of papers makes a
% one-size-fits-all approach a daunting prospect. For instance, compsoc 
% journal papers have the author affiliations above the "Manuscript
% received ..."  text while in non-compsoc journals this is reversed. Sigh.

\author{Jielong~Yang,~\IEEEmembership{Student Member,~IEEE,}			
	and Wee~Peng~Tay,~\IEEEmembership{Senior~Member,~IEEE}
	% <-this % stops a space
	\thanks{This work was supported in part by the Singapore Ministry of Education Academic Research Fund Tier 2 grant MOE2018-T2-2-019 and by A*STAR under its RIE2020 Advanced Manufacturing and Engineering (AME) Industry Alignment Fund – Pre Positioning (IAF-PP) (Grant No. A19D6a0053).}
	\thanks{J. Yang is with the School of Artificial Intelligence, Jilin University, China. Email: jyang022@e.ntu.edu.sg.}		
	\thanks{W. P. Tay is with the School of Electrical and Electronic
		Engineering, Nanyang Technological University, Singapore. Email:		wptay@ntu.edu.sg.}% <-this % stops a space
}

\IEEEtitleabstractindextext{%
\begin{abstract}
	The problem of estimating event truths from conflicting agent opinions in a social network is investigated. An autoencoder learns the complex relationships between event truths, agent reliabilities and agent observations. A Bayesian network model is proposed to guide the learning process by modeling the relationship of the autoencoder's outputs with different variables. At the same time, it also models the social relationships between agents in the network. The proposed approach is unsupervised and is applicable when ground truth labels of events are unavailable. A variational inference method is used to jointly estimate the hidden variables in the Bayesian network and the parameters in the autoencoder. Experiments on three real datasets demonstrate that our proposed approach is competitive with, and in most cases better than, several state-of-the-art benchmark methods. 
\end{abstract}

% Note that keywords are not normally used for peerreview papers.
}

% make the title area
\maketitle

% To allow for easy dual compilation without having to reenter the
% abstract/keywords data, the \IEEEtitleabstractindextext text will
% not be used in maketitle, but will appear (i.e., to be "transported")
% here as \IEEEdisplaynontitleabstractindextext when the compsoc 
% or transmag modes are not selected <OR> if conference mode is selected 
% - because all conference papers position the abstract like regular
% papers do.
\IEEEdisplaynontitleabstractindextext
% \IEEEdisplaynontitleabstractindextext has no effect when using
% compsoc or transmag under a non-conference mode.

% For peer review papers, you can put extra information on the cover
% page as needed:
% \ifCLASSOPTIONpeerreview
% \begin{center} \bfseries EDICS Category: 3-BBND \end{center}
% \fi
%
% For peerreview papers, this IEEEtran command inserts a page break and
% creates the second title. It will be ignored for other modes.
\IEEEpeerreviewmaketitle

\IEEEraisesectionheading{\section{Introduction}\label{sec:introduction}}
% Computer Society journal (but not conference!) papers do something unusual
% with the very first section heading (almost always called "Introduction").
% They place it ABOVE the main text! IEEEtran.cls does not automatically do
% this for you, but you can achieve this effect with the provided
% \IEEEraisesectionheading{} command. Note the need to keep any \label that
% is to refer to the section immediately after \section in the above as
% \IEEEraisesectionheading puts \section within a raised box.

\IEEEPARstart{I}{t} is common for agents in a social network to report conflicting opinions \cite{karger2013efficient,AceDahLobOzd:11,HoTayQue:J15,Tay:J15,GraSurAli2016,HuaWan2016,KanTay2019,BerBor2015}. Some of the agents are unreliable and maybe biased. The majority voting method fuses the agents' opinions together by treating the opinions from a majority of agents as the estimated truth. This is based on the assumption that all agents have the same reliability \cite{LiGaoMen2016}. This assumption may not be reasonable when agents are from different backgrounds and their reliabilities or biases vary widely. Truth discovery methods have been proposed to estimate event truths in consideration of agent reliabilities. Crowdsourcing \cite{CheLiaChe2016,TonCheZho2018,TiaZhoZhu2018,AlsAmeGau2017} can be regarded as an application of truth discovery. 

The relationships between event truths, agent reliabilities and agent observations are complex. To model these relationships, various assumptions of agent reliabilities are adopted in the literature. 
%<*new-ref-1>
Different from the method proposed in \cite{WanMaJin2018} that uses both textual and visual features to detect event states, we only consider the event observations in this paper.
%</new-ref-1>
The papers \cite{WanKapLe2012, ZhaRubGem2012, WanAmiLi2014, HuaWan2016, YaoHuLi2016} developed probabilistic models for truth discovery from binary observations with an agent's reliability being the probability an event is true given that the agent reports it to be true. Multi-ary observations are considered in \cite{YinHanPhi2008, DonBerSri2009, LiLiGao2015a, ZhaWanZha2017}. A Bayesian method named TruthFinder was proposed by \cite{YinHanPhi2008} to iteratively estimate the probability that each agent is correct and the event truths.  In \cite{DonBerSri2009}, a Bayesian method named AccuSim was developed to learn if agents copy opinions from others, their reliabilities and the event truths. The CATD method was proposed in \cite{ LiLiGao2015a} for the case where most agents provide only limited opinions. In \cite{ZhaWanZha2017}, a hidden Markov model is used to infer event truths and agent reliabilities that evolve over time, and \cite{ZhaWuHua2017} used a maximum likelihood estimation approach to estimate event truths when each agent's reliability may vary across events. In \cite{LiRubCoh2019}, the authors considered the case where each event is observed by many agents (in crowdsourcing applications). In this case, the majority voting method achieves good performance. The authors proposed a graphical model that outperforms majority voting. However, their approach is not applicable when each event is observed by very few agents, e.g., in social learning applications \cite{AceDahLobOzd:11,HoTayQue:J15,Tay:J15,HuaWan2016}. 
%<*new-ref-2>
In \cite{BroGotKas2017}, a Restricted Boltzmann Machine (RBM) based truth discovery method is proposed based on consideration of effectiveness, efficiency and robustness. In \cite{WanCheKap2016}, the authors proposed an uncertainty aware approach called Kernel Density Estimation from Multiple Sources (KDEm) to estimate the probability distributions of the trustworthy opinions.
%</new-ref-2>
Note that all these methods do not make use of the relationships between agents to aid the inference process.

In \cite{KimGha2012}, the authors proposed a method called Bayesian Classifier Combination (BCC) for truth discovery using a confusion matrix to represent the reliability of each agent. The use of confusion matrices generally outperforms models that use scalar agent reliabilities, as demonstrated by \cite{ZheLiLi2017}. In this paper, BCC was shown to be amongst the best methods in decision making and single-label tasks. In the BCC method, it is difficult to infer accurately an agent's reliability if it observes only a small subset of events. To mitigate this problem, the Community BCC (CBCC) model was proposed by \cite{VenGuiKaz2014}, which grouped agents with similar backgrounds into communities and assumed that the confusion matrix of an agent is a perturbation of the confusion matrix of its community. In \cite{LyuOuyWan2019}, the co-occurrence of two agents across different events is considered but communities of agents are not considered in this paper. In \cite{LiRubCoh2019a}, the agent correlation is considered by estimating agent reliabilities at sub-type levels instead of event classes. Direct agent relationships are however still not considered. 

The papers \cite{GraSurAli2016} and \cite{YinGraSur2016} showed that agents in a crowd are related through social ties and are influenced by each other. 
%<*tag:compare-with-previous-1>
In \cite{MaTayXia:J18}, the authors adopted a model in which an agent can be influenced by another agent to change its observation to match that of the influencer, while \cite{WanAmiLi2014} assumed that agents' dependency graphs are disjoint trees. This was extended to general dependency graphs in \cite{HuaWan2016,YaoHuLi2016}. In our previous work \cite{YanWanTay2019}, we considered the use of social network information and community detection to aid in truth discovery. We called our approach VISIT. Note that in these methods, the relationships among event truths, agent reliabilities and agent observations are  modeled by predefined models. Such assumptions limit the flexibility of the model to fit complex real data.
%</tag:compare-with-previous-1>

Neural networks have shown good promise in modeling nonlinear relationships in many applications \cite{AleIlyGeo2012, Jef1990}.
%<*new-ref-3>
In \cite{MarArgWan2017}, the relationship between source reliability and claim truthfulness is modeled using a multi-layer neural network and achieved promising accuracy for truth discovery. In \cite{LiQinRen2017}, a memory network based method is used to model the non-linear relationship between source reliability and claim truthfulness. However, these two methods are both supervised learning methods that do not consider the interpretable structures of agents (e.g., communities of agents).
%</new-ref-3> 
We believe the application of neural networks are more effective in the truth discovery problem if the following issues are properly resolved:
\begin{enumerate}[(a)]	
	\item \emph{Unsupervised learning:} For the truth discovery problem, it is often difficult to obtain enough ground truth labels of events to learn a supervised model. Thus, developing an unsupervised model is important and meaningful. 
	\item \emph{Modeling interpretable structures:} In the truth discovery problem, the observations from agents often imply hidden structures. For instance, agents having similar background, culture, and socio-economic standing may form communities and share similar reliabilities\cite{VenGuiKaz2014,YanWanTay2019}. Successfully discovering the hidden structures can improve the performance of the truth discovery model. However, neural networks are not good at modeling interpretable structures\cite{JohDuvWil2016}. 
	\item \emph{Dealing with dependency:} The interpretable structures in item (b) result in different data samples to have dependencies, which violates the independence assumption that is used in many neural network based methods\cite{DunKriBi2007}.
\end{enumerate}

In this paper, to solve the first issue, an unsupervised deep autoencoder is used to learn the complex relationship among event truths, agent reliabilities and agent observations. Autoencoders \cite{KinWel2013} are a kind of unsupervised artificial neural network widely used to learn data features. However, the optimization process of an autoencoder is easily stuck in less attractive local optima \cite{ZonSonMin2018}, thus proper model constraints are required to obtain better performance. The constraints are introduced by Bayesian networks in our model. Bayesian network models provide a natural way to characterize the relationship among variables in an unsupervised way \cite{AirBleFie2008,TehJorBea2012,BleNgJor2003}. In this paper, a Bayesian network model is proposed to model the social relationships between agents, which we assume affect each agent's reliability, and to further constrain the learning of the autoencoder. 
%<*tag:compare-with-previous-2>
Our approach combines the strengths of unsupervised learning in modeling nonlinear relationships and the strengths of Bayesian networks in characterizing hidden interpretable structures.
%</tag:compare-with-previous-2>
Our model is not a straightforward concatenation of a Bayesian network and an autoencoder, but constructs a network of the autoencoder's output variables. The Bayesian network and the autoencoder are learned iteratively.

The rest of this paper is organized as follows. In \cref{sec:model}, we present our model assumptions. In \cref{sec:VarInfer}, we present our proposed variational inference approach, which is iterative in nature. We show how each update step is derived. Experimental results are discussed in \cref{sec:ExperimentResults} and we conclude in \cref{sec:conclusion}.

\emph{Notations:} We use boldfaced characters to represent vectors and matrices. Suppose that $\bM$ is a matrix, then $\bM(m,\cdot)$, $\bM(\cdot,m)$, and $\bM(m,n)$ denote its $m$-th row, $m$-th column, and $(m,n)$-th element, respectively. The vector $(x_1,\ldots,x_N)$ is abbreviated as $(x_i)_{i=1}^N$ or $(x_i)$ if the index set that $i$ runs over is clear from the context. The $i$-th element of a vector $\bx$ is $\bx(i)$. Let $\bone$ denote a column vector of all 1's and $\bI$ be the identity matrix. The vectorized version of $\bM$ with the columns of $\bM$ stacked together as a single column vector is denoted as $\vect(\bM)$. We use $\Cat{p_1,\ldots,p_K}$ and $\N{\bU}{\bV}$ to represent the categorical distribution with category probabilities $p_1,\ldots,p_K$ and the normal distribution with mean $\bU$ and covariance $\bV$, respectively. The notation $\sim$ means equality in distribution. The notation $y\mid x$ denotes a random variable $y$ conditioned on $x$, and $p(y\mid x)$ denotes its conditional probability density function. $\E$ is the expectation operator and $\E_q$ is expectation with respect to the probability distribution $q$. The notation $\mathrm{prob}(\theta)$ for a random variable $\theta$ is a column vector whose $r$-th value is the probability $\P(\theta=r)$. We use $I(a,b)$ to denote the indicator function, which equals 1 if $a=b$ and 0 otherwise. We use $|\calS|$ to represent the cardinality of the set $\calS$.

%An autoencoder is used to learn the relationship between reliabilities of agents, true states of events and the opinions of agents. Moreover, a Bayesian network model is used to introduce constraints to the autoencoder and at the same time incorporate the community information of social network into the autoencoder. 
\renewcommand{\arraystretch}{1.8}
\begin{table*}[!tb]
	\caption{Summary of commonly-used symbols.} % title of Table
	\centering % used for centering table
	\begin{tabular}{L{3cm} L{5cm} C{5cm}} % centered columns (4 columns)		
		\hline %inserts double horizontal lines
		Symbol & Description & Variational Parameter in Section \ref{sec:VarInfer} \\ [0.5ex] % inserts table
		%heading
		\hline
		$\bM=(\bM(n,j))_{\substack{1\leq n \leq N,\\ 1\leq j\leq J}}$ & $\bM(n,j)$ is the set of observations of agent $n$ about event $j$.  & N.A.\\
		\hline
		$\bC=(\bC_n)_{n=1}^N$ & $\bC_n$ is the $R_1\times R_2$ reliability matrix of agent $n$. & Neural network parameters $\bw_R$ of the reliability encoder network.\\
		\hline
		$\btheta=(\theta_j)_{j=1}^J$	& $\theta_j$ is the true state of event $j$. & Neural network parameters $\bw_E$ of the event encoder network.\\
		\hline
		\baselineskip=15pt		
		$\bo=(\bo_n)_{n=1}^N$, $\bu=(\bu_j)_{j=1}^{J}$, $\bd=(\bd_{n,j})_{\substack{1\leq n \leq N,\\ 1\leq j\leq J}}$ &	$\bo_n,\bu_j,\bd_{i,j}$ are the outputs of the reliability encoder network, the event encoder network and the decoder network, respectively. & N.A.\\
		\hline
		$\tbC=(\tbC_k)_{k=1}^K$ & $\tbC_k$ is the reliability matrix of community $k$. & $(\tbmu_k, \tbsigma_k)_{k=1}^K$ \\
		\hline % inserts single horizontal line	
		$\bA(n,m)=1\text{ (or 0)}$ for $n,m\in\{1,...,N\}$     &There is a (or no) social connection between agents $n$ and $m$. & N.A.\\ % [1ex] adds vertical space
		\hline %inserts single line
		$\bz=(z_{n\rightarrow m})_{\substack{1\leq n \leq N,\\ 1\leq m\leq N, m\neq n}}$ & $z_{n\rightarrow m}$ is the index of the community agent $n$ subscribes to under the social influence of agent $m$.& $\bphi=(\phi_{n\rightarrow m,k})_{\substack{1\leq k\leq K,\\1\leq n \leq N,\\ 1\leq m\leq N, m\neq n}}$\\
		\hline
		$\bbeta=(\beta_k)_{k=1}^K$ & $\beta_k$ is the social network parameter defined in \eqref{eq:prior_Dnm}. & $\blambda=(\blambda_k)_{k=1}^K$; ${\blambda_k=(G_k,H_k)}$\\
		\hline
		$\bpi=(\bpi_{n})_{n=1}^N=(\pi_{n,k})_{\substack{1\leq n \leq N,\\ 1\leq k\leq K}}$ & $\bpi_n$ is the distribution of $s_n$ and $z_{n\rightarrow m}$, which are defined in \eqref{eq:prior_s} and \eqref{eq:prior_znm}.& $\bgamma=(\gamma_{n,k})_{\substack{1\leq n \leq N,\\ 1\leq k\leq K}}$\\
		\hline
		$\bs=(s_n)_{n=1}^N$ & $s_n$ is the community index of agent $n$.& $\bpsi=(\psi_{n,k})_{\substack{1\leq n \leq N,\\ 1\leq k\leq K}}$\\
		\hline	
		\baselineskip=15pt
		$(\bU_k,\bV_k)$, $\balpha$, $(g_0,h_0)$, $\bp^{MV}=(\bp^{MV}_j)_{j=1}^J$, $\epsilon$, $b$, $\bb'$,
		$\tau$  &  Hyper-parameters defined in \eqref{eq:prior_tbC_k}, \eqref{eq:define_alpha}, \eqref{eq:prior_znm}, \eqref{eq: define_w_MV}, \eqref{eq:prior_Dnm}, \eqref{eq:bC_given_b}, \eqref{eq:bC_b'}, and \eqref{eq:theta_tao},   respectively.  & N.A.\\	
		\hline	
	\end{tabular}
	\label{table:Notations} % is used to refer this table in the text
\end{table*}

\section{System Model}\label{sec:model}

Suppose that $N$ agents in a social network with different reliabilities observe $J$ events and each event $j=1,\ldots,J$ has a state $\theta_j$, which can take $R$ possible states. We consider the problem of estimating event truths or states from conflicted agent opinions when each agent only observes a subset of the events. The symbols used in this paper are summarized in \cref{table:Notations}. 

%<*tag:generalized-M>
In our model, the observation matrix $\bM$ is an $N \times J$ generalized matrix where each element $\bM(n,j)$ represents the set of observations of agent $n$ about event $j$. Each agent may form multiple observations about an event. This is useful to model the case where an agent may be undecided about the state of the same event. An entry $\bM(n,j)$ is null or the empty set if the agent $n$ does not observe event $j$.
%</tag:generalized-M>
For non-null $\bM(n,j)$, we assume that it is generated from $\bC_n$, which represents the reliability of agent $n$'s opinion about the ground truth state $\theta_j$ of event $j$. The reliability matrix $\bC_n$ is a $R_1 \times R_2$ matrix, where $R_1$ and $R_2$ are two hyper parameters. In this paper, we use a general matrix to represent an agent's reliability, and this includes both the reliability concepts of \cite{YinHanPhi2008, DonBerSri2009, LiLiGao2015a, ZhaWanZha2017, ZhaRubGem2012} and confusion matrix of \cite{KimGha2012,ZheLiLi2017,VenGuiKaz2014}. We also assume that a social network connecting the agents is known and its graph adjacency matrix is given by $\bA$, where $\bA(n,m)=1$ iff agent $n$ and agent $m$ are connected. Our target is to estimate $\btheta\triangleq(\theta_j)_{j=1}^J$ from $\bM$ and $\bA$. 

\subsection{Observation Model}\label{subsec:observation_model}

Let $\bC=(\bC_n)_{n=1}^N$. The relationship between $\bM$ and $(\bC,\btheta)$ is complex and nonlinear, and thus it is challenging to find a proper analytical model for it. Moreover, a data set with a large number of accurate labels is usually unavailable, which hinder the application of supervised learning in the truth discovery problem. To solve these two issues, we model the relationship between $(\bC,\btheta)$ and $\bM$ with a multi-layer neural network and perform inference using an unsupervised autoencoder in \cref{sec:VarInfer}.

We represent the observation model for $\bM$ by a neural network that learns to decode the agent reliabilities and event states back to the observations. The input to the observation model is $(\bC_n,\theta_j)$ for $n=1,\ldots,N$ and $j=1,\ldots,J$. We calculate $(\bw_{D_1}\mathrm{prob}(\theta_j))\odot\vect(\bC_n)$, where $\odot$ represents element-wise multiplication and $\bw_{D_1}$ is a $R_1R_2\times R$ learnable matrix.  Next, we input the obtained result to multiple fully connected layers. We denote all the learnable parameters of the observation model including $\bw_{D_1}$ as $\bw_D$. From the output layer of the observation model, we obtain an $R \times 1$ vector $\bd_{n,j}$. 
%<*tag:why-bernoulli>	
For all $r \in [1,R]$, we assume 
\begin{align}\label{MBern}
\onezero{\bM(n,j)}(r) \sim \Bern{\bd_{n,j}(r)},
\end{align}
where $\onezero{\cdot}$ is the sum of the one-hot vector representation of the elements of its set argument, and $\Bern{\cdot}$ denotes the Bernoulli distribution. We use a Bernoulli distribution to model the case where each agent is may be unsure of its observations and has multiple observations about an event. 
%</tag:why-bernoulli>

\subsection{Bayesian Model Constraints}

\begin{figure}[!htb]
	\centering
	\includegraphics[width=7cm]{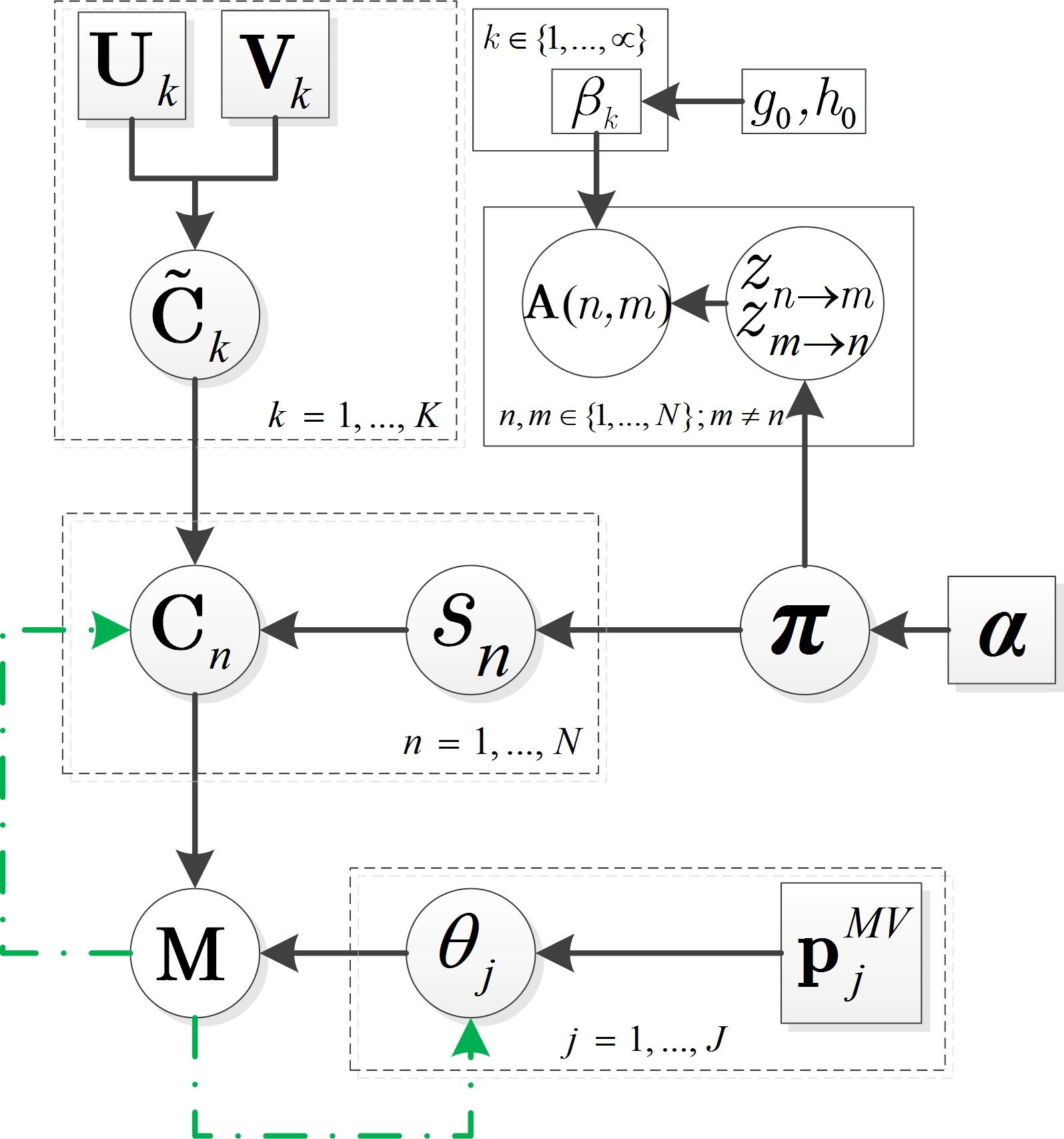}
	\caption{Our proposed Bayesian network model. The arrows from the nodes labeled as $\bC_n$ and $\btheta_j$ to $\bM$ represent the observation model or decoder network in \cref{fig:autoencodermodel}. The dotted arrows from $\bM$ to the nodes labeled as $\bC_n$ and $\btheta_j$ represent the reliability encoder and event encoder in \cref{fig:autoencodermodel}, respectively. The dotted arrows do not form part of our Bayesian network model, but are instead used in the variational inference of our model in \cref{sec:VarInfer}.}
	\label{fig:model}	
\end{figure}

One problem of learning the observation model is that optimizing its neural network weights can become stuck in less attractive local optima \cite{ZonSonMin2018}. To mitigate this, proper constraints on key latent variables $\bC$ and $\btheta$ need to be introduced. In this paper, we use a Bayesian network model (see \cref{fig:model}) to construct interpretable constraints. The Bayesian network model not only guides the learning process but also enables to use the community information of the social network linking the agents together. We explain each component of our Bayesian network model below.

\subsubsection{Community reliability matrix \texorpdfstring{$\tbC_n$}{C}}

We assume that a social network connecting the $N$ agents is known. Agents in a social network tend to form communities \cite{ForHri2016}, where a community consists of agents with similar opinion. The community that an agent belongs to is stochastic and unknown \emph{a priori} but the maximum possible number of communities in the network is known to be $K$. We can thus assign an index $1,2,\ldots,K$ in an arbitrary fashion to each community. Let $s_n$ be the community index of agent $n$. We discuss the probability model governing $s_n$ below. Here, we describe how $\bC_n$ depends on $s_n$. For each $k=1,\ldots,K$, let $\tbC_k$ be a matrix of the same size as $\bC_n$ representing the reliability of community $k$. 
%<*tag:why-perturbed>
We assume 
\begin{align}
\bC_n(r_1,\cdot)\mid \braces*{\tbC_k(r_1,\cdot), s_n=k} \sim \N{\tbC_{k}(r_1,\cdot)}{b'\bI}, \label{eq:bC_b'}
\end{align}
for $r_1 \in [1,R_1]$, where $b'$ is a hyperparameter. Then $\bC_n$ can be regarded as a perturbed version of $\tbC_{s_n}$ since $\E[\bC_n(r_1,\cdot) \mid \tbC_k(r_1,\cdot), s_n=k]=\tbC_{k}(r_1,\cdot)$ for $r_1 \in [1,R_1]$.
%</tag:why-perturbed> 
We assume that for $k=1,\ldots, K$, and $r_1 \in [1,R_1]$,
\begin{align}
\tbC_k(r_1,\cdot)\sim \N{\bU_k(r_1,\cdot)}{\bV_k},\label{eq:prior_tbC_k}
\end{align}
where $\bU_k$ and $\bV_k$ are hyper parameters. 
%<*tag:why-normal>
Here we use the normal distribution as the prior as it is an effective prior in many real applications and it allows us to simplify the subsequent inference equations.
%</tag:why-normal>
From \cref{fig:autoencodermodel}, we see that $\{\bC_n\}$ are learned from different inputs of the reliability encoder. Different from a traditional autoencoder \cite{BenYaoAla2013,GooBenCou2016}, $\{\bC_n\}$ are not independent and their relationship is modeled by the Bayesian network in \cref{fig:model}.  The Bayesian network model guides the learning process of the autoencoder. 

\subsubsection{Community index \texorpdfstring{$s_n$}{s}}
We model the community index $s_n$ of agent $n$ as 
\begin{align}
s_n\sim \Cat{\bpi_n}, \label{eq:prior_s}
\end{align}
where the mixture weights
\begin{align} 
\bpi_n = (\pi_{n,k})_{k=1}^K \sim \Dir{\balpha}, \label{eq:define_alpha}
\end{align}
with $\balpha$ being a concentration hyperparameter and $\Dir{\balpha}$ is the  Dirichlet distribution.
%<*tag:why-Dirichlet> 
Here we use the Dirichlet distribution since the support of a Dirichlet distribution can be regarded as the probabilities of categorical events. Besides, the Dirichlet distribution is the conjugate prior distribution of the categorical distribution in \cref{eq:prior_s} and \cref{eq:prior_znm}, which thus allows us to derive analytically the posterior distribution of $\bpi_{n}$.
%</tag:why-Dirichlet>
We use the mixed membership stochastic block model (MMSB)\cite{AirBleFie2008} to model the social connection $\bA(n,m)$ between agents $n$ and $m$. In this model, $z_{n\rightarrow m}$ is the community whose belief agent $n$ subscribes to due to the social influence from agent $m$. Under the influence of different agents, agent $n$ may subscribe to the beliefs of different communities. If both agents $n$ and $m$ subscribe to the belief of the same community, they are more likely to be connected in the social network. We assume the following:
\begin{align}
z_{n\rightarrow m}\mid \bpi_n& \sim \Cat{ \bpi_n },\nonumber\\
z_{m\rightarrow n}\mid \bpi_m& \sim \Cat{ \bpi_m }, \nonumber\\
\beta_{k}& \sim \Beta{g_0}{h_0}, \label{eq:prior_znm}
\end{align}
where $\Beta{g_0}{h_0}$ is the beta distribution with parameters $g_0, h_0 >0$, $k=1,\ldots,K$, and
\begin{align}
&\P(\bA(n,m)=1\mid z_{n\rightarrow m},z_{m\rightarrow n},\beta_{z_{n\rightarrow m}}) \nonumber\\
&=
\begin{cases}
\beta_{z_{n\rightarrow m}}, & \text{if}\ z_{n\rightarrow m}=z_{m\rightarrow n}, \\	
\epsilon, & \text{if}\ z_{n\rightarrow m}\ne z_{m\rightarrow n},
\end{cases}
\label{eq:prior_Dnm}
\end{align}
with $\epsilon$ being a small constant. 
%<*tag:why-beta>
In \cref{eq:prior_znm}, we use beta distribution since the beta distribution is a conjugate prior for the Bernoulli distribution in \cref{eq:prior_Dnm}.  Note  that $\bA$ is independent of $\bpi$ when $\bz$ is given, as shown in \cref{fig:model}.
%</tag:why-beta>
\subsubsection{Event states \texorpdfstring{$\btheta$}{theta}}

A direct method to perform truth discovery is majority voting, i.e., selecting the opinion expressed by the most number of agents as the true state of the event. This assumes that all agents have the same reliability, and that agents are more likely to give the correct opinion than not. Without any prior information, this is a reasonable assumption. Therefore, we let the prior of $\btheta$ to be given by
\begin{align}
\theta_j\sim \Cat{\bp_j^{MV}} \label{eq: define_w_MV}
\end{align}
for each $j=1,\ldots,J$, where $\bp_j^{MV}(r)$ for $r=1,\ldots,R$ is the proportion of agents who thinks that the state of event $j$ is $r$. We assume that $\{\theta_j: j=1,\ldots,J\}$ are independent.

\section{ART: Autoencoder Truth Discovery}\label{sec:VarInfer}

In this section, we propose an autoencoder based on unsupervised variational inference \cite{HofBleWan2013} for the Bayesian model in \cref{fig:model}. 

Let $\bbeta=(\beta_k)$, $\bz=(z_{n\rightarrow m})$, $\bs=(s_n)$, $\bpi=(\bpi_{n})$, $\btheta=(\theta_j)$, and $\tbC=(\tbC_{k})$. For simplicity, let $\bOmega\triangleq (\bbeta, \bz,  \bpi, \bs, \tbC, \bC, \btheta)$.  As the closed-form of the posterior distribution $p(\bOmega\mid \bM,\bA)$ is not available, the variational inference method uses a proposal or variational distribution $q(\bOmega; \bLambda)$ to approximate the posterior distribution, where the parameters in the vector $\bLambda$ are called the variational parameters. Note that $\bM$ and $\bA$ are assumed to be observed throughout and not included explicitly in our notation for $q$. More specifically, the variational parameters are selected to minimize the following cost function:
\begin{align}
\calL=-\E{q}[\log p(\bOmega\mid \bM,\bA) -\log q(\bOmega; \bLambda)],\label{eq:costfunction}
\end{align}  
where the expectation is over the random variable $\bOmega$ with distribution $q(\bOmega;\bLambda)$ conditioned on $\bM$ and $\bA$. To simplify the optimization procedure, we use the mean-field assumption that is widely used in the literature\cite{BleJor2006,HofBleWan2013} by choosing
\begin{dmath}
	q(\bOmega; \bLambda)= q(\bbeta; \blambda) q(\bz; \bphi) q(\bpi; \bgamma) q(\bs; \bpsi)q(\tbC;\tbmu,\tbsigma) q(\bC; \bw_R)q(\btheta; \bw_E) \label{eq:meanfield_q},
\end{dmath}
where $\bLambda=(\blambda,\bphi,\bgamma,\bpsi,\tbmu,\tbsigma,\bw_R,\bw_E)$, $\blambda=(\blambda_k)_{k=1}^K$, $\bphi=(\phi_{n,m})_{n,m}$, $\bgamma=(\gamma_{n,k})_{n,k}$, $\bpsi=(\psi_{n,k})_{n,k}$, $\tbmu=(\tbmu_k)_{k=1}^K$, and $\tbsigma=(\tbsigma_k)_{k=1}^K$ are the variational parameters. 

In the sequel, to simplify notations, we omit the variational parameters in our notations, e.g., we write $q(\bbeta)$ instead of $q(\bbeta;\blambda)$. We let $q(\bC,\btheta)=q(\bC)q(\btheta)$. From the graphical model in \cref{fig:model}, we obtain
\begin{dmath}
	{p(\bOmega \mid \bM,\bA)} \propto {p(\tbC)}{p(\bs \mid \bpi)}{p(\bz\mid\bpi)}{p(\bpi)}{p(\bA\mid\bbeta,\bz)}{p(\bbeta)}{p(\bC \mid \tbC,\bs)}\\\quad\cdot{p(\bM\mid\bC,\btheta; \bw_D)}{p(\btheta)},\label{eq:p_phi_M_D}
\end{dmath}
where $p(\bM\mid\bC,\btheta; \bw_D)$ is the conditional distribution of the output of the decoder network in \cref{fig:autoencodermodel}. We have made its dependence on the decoder network parameters $\bw_D$ explicit.

To find the variational parameters, we perform an iterative optimization of $\calL$ in which the optimal parameter solutions are updated iteratively at each step. 
%We show later in this section that it is possible to find analytical update rules for the variational parameters $\blambda,\bphi,\bgamma,\bpsi,\tbmu,\tbsigma$. However, finding analytical forms of the update rule of $(\bw_R, \bw_E)$ is more challenging. 
We substitute \eqref{eq:meanfield_q} and \eqref{eq:p_phi_M_D} into \eqref{eq:costfunction} to obtain
\begin{dmath}
	\calL=\calL_1 + \calL_2 +\text{constant},
\end{dmath}
where the constant term does not contain any variational parameters, and
\begin{dgroup*}
	\begin{dmath}
		\calL_1\triangleq-\E{q(\bC,\btheta,\tbC,\bs)}\left[{\log p(\bC \mid \tbC,\bs)}+\log {p(\bM\mid\bC,\btheta; \bw_D)}+\log p(\btheta)-\log {q(\bC,\btheta )}\right],\label{L1}
	\end{dmath}
	\begin{dmath}
		\calL_2\triangleq-\E{q(\tbC,\bs,\bpi,\bz,\bbeta )}\left[\log p(\tbC)+{\log p(\bs \mid \bpi)}+{\log p(\bz\mid\bpi)}+\log p(\bpi)+{\log p(\bA\mid\bbeta,\bz)}+\log p(\bbeta) -\log q(\tbC)-\log q(\bbeta)-\log q(\bz)-\log q(\bpi)-\log q(\bs)\right]. 
	\end{dmath}
\end{dgroup*}
We update $q(\bC)$, $q(\btheta)$ by minimizing $\calL_1$ and update $q(\bbeta)$, $ q(\bz)$, $ q(\bpi)$ by minimizing $\calL_2$. Furthermore, we update $q(\tbC)$ and $q(\bs)$ by minimizing 
\begin{dgroup*}
	\begin{dmath*}
		\calL_3 \triangleq -\E{q(\bC,\tbC,\bs,\bpi)}\left[{\log p(\bC \mid \tbC,\bs)}+\log p(\tbC)+{\log p(\bs \mid \bpi)}-\log q(\tbC) -\log q(\bs)\right]
	\end{dmath*} 
	\begin{dmath}\label{eq:L3_s}
		=-\E{q(\bC,\tbC,\bs,\bpi)}\left[{\log p(\bs\mid \bC,\tbC,\bpi)}+{\log p(\bC\mid\tbC)}+{\log p(\tbC)}-{\log q(\tbC)} -{\log q(\bs)}\right]
	\end{dmath}
	\begin{dmath}\label{eq:L3_C}
		=-\E{q(\bC,\tbC,\bs,\bpi)}\left[{\log p(\tbC\mid \bC,\bs)}+{\log p(\bC\mid\bs)}+{\log p(\bs \mid \bpi)}-{\log q(\tbC)} -{\log q(\bs)}\right].
	\end{dmath} 
\end{dgroup*}
Equations \eqref{eq:L3_s} and \eqref{eq:L3_C} show $q(\bs)$ and $q(\tbC)$ can be updated by minimizing $$-\E{q(\bC,\tbC,\bs,\bpi)}\left[{\log p(\bs\mid \bC,\tbC,\bpi)}-{\log q(\bs)}\right]$$ and  $$-\E{q(\bC,\tbC,\bs,\bpi)}\left[{\log p(\tbC\mid \bC,\bs)}-{\log q(\tbC)}\right],$$ respectively.

The variational parameters are optimized and the variational distributions updated iteratively in a procedure that we call AutoencodeR Truth (ART) discovery (since we make use of an autoencoder network described below). Its high-level pseudo code for the $i$-th iteration is shown in \cref{alg:inferenceAlgorithm}. In the following, we describe how the variational distributions are chosen and how the estimate for the optimal variational parameters are updated in each iteration.

\begin{algorithm}[!htb]
	\caption{ART ($i$-th iteration)}\label{alg:inferenceAlgorithm}
	\begin{algorithmic}
		\renewcommand{\algorithmicrequire}{\textbf{Input:}}
		\renewcommand{\algorithmicensure}{\textbf{Output:}}
		\REQUIRE Variational parameters in $(i-1)$-th iteration, opinions $\bM$, social network data $\bA$.
		\ENSURE  Variational parameters in $i$-th iteration. 		
		\FOR {each agent $n$ in $\{1, \ldots,N\}$}		
		\FOR {each agent pair $(n,m)$ in $\{(n,m)\}_{m=1,m\neq n}^N$}
		\STATE Update $\bphi_{n\rightarrow m}$ and $\bphi_{m \rightarrow n }$ using \eqref{eq:update_phi_D1} and \eqref{eq:update_phi_D0}.		
		\ENDFOR		
		\STATE Update $\bpsi_n$ using \eqref{eq:update_psi}. 
		\STATE Update $\bgamma_n$ using \eqref{eq:update_gamma}.
		\STATE Sample $\bC_n$ using \eqref{eq:sample_bC'}.	
		\STATE Sample $s_n$  from $q(s_n)=\Cat{(\psi_{n,k})_{k=1}^{K}}$.	 
		\ENDFOR
		\STATE Update $\blambda$ using \eqref{eq:update_Gk} and \eqref{eq:update_Hk}.	 
		\STATE Update $\tbmu$ and $\tbsigma$ using \eqref{eq:update_tmu} and \eqref{eq:update_tsigma}.		
		\STATE Sample $\tbC$, and $\btheta$ using \eqref{eq:q_tbC} and \eqref{eq:sample_hbtheta}.
		\STATE Learn the autoencoder (i.e., update $\bw_R$, $\bw_E$, and $\bw_D$ in Section \ref{eq:Section_update_autoencoder}).
		\RETURN $\bphi$, $\bpsi$, $\bgamma$, $\bC$, $\blambda$, $\tbmu$, $\tbsigma$, $\tbC$, and $\btheta$.	
	\end{algorithmic} 
\end{algorithm}

\subsection{Reliability and Event Encoders}\label{subsec:encoders}

\begin{figure*}[!htb]
	\centering
	\includegraphics[width=0.8\textwidth]{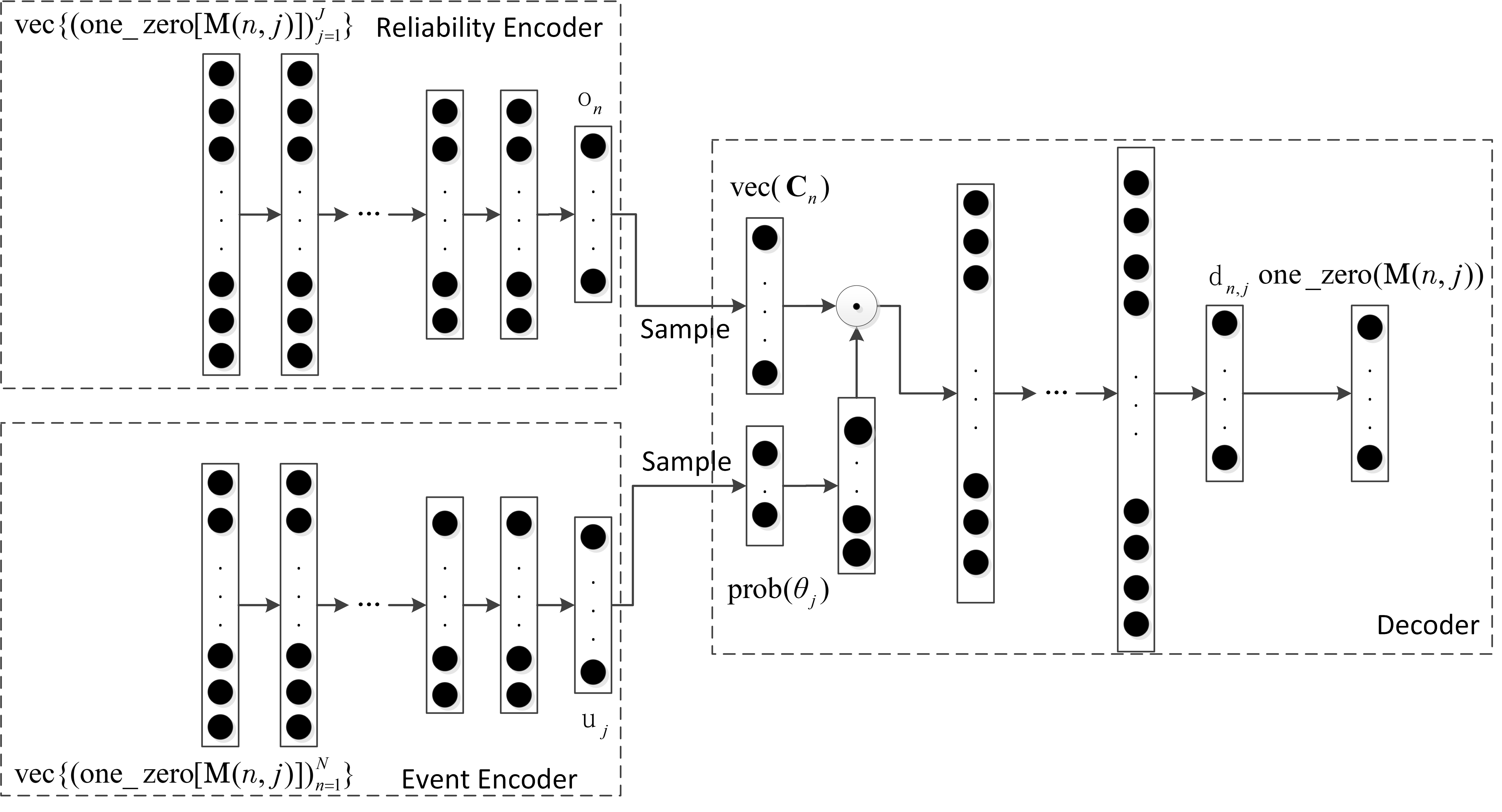}
	\caption{Autoencoder model. }
	\label{fig:autoencodermodel}	
\end{figure*}

We model $q(\bC; \bw_R)$ and $q(\btheta; \bw_E)$ as two encoder networks with parameters $\bw_R$ and $\bw_E$, respectively. These two encoders form part of the autoencoder shown in \cref{fig:autoencodermodel} with the observation model for $\bM$ given in \cref{subsec:observation_model} as the decoder network. The details of the two encoder networks are as follows.
\begin{enumerate}[(a)]
	\item \label{sec:Encoder1} 
	The reliability encoder network is used to infer the reliability of each agent $n$, with $\onezero{\bM(n,\cdot)}$ as its input.  If $\bM(n,j)$ is null, then $\onezero{\bM(n,j)}$ is a zero vector.  All the layers of this encoder are fully connected. Let its parameters be $\bw_R$ and its $R_1R_2\times 1$ softmax output be $\bo_n$. We assume 
	\begin{align}
	\vect(\bC_n)(i)\sim \N{\bo_n(i)}{b}, \label{eq:bC_given_b}
	\end{align}
	where $b$ is a hyperparameter. 
	
	\item \label{sec:Encoder2}
	The event encoder network is used to infer the state of each event from $\bM$. All the layers are also fully connected. Let its parameters be $\bw_E$ and its output be $\bu_j$. We assume
	\begin{align}
	\theta_j\sim\Cat{\bu_j}.\label{eq:hbtheta_j}
	\end{align}
\end{enumerate}

We let $q(\bC; \bw_R)$ correspond to the Gaussian distribution \cref{eq:bC_given_b}, and $q(\btheta; \bw_E)$ correspond to the distribution \cref{eq:hbtheta_j}.

\subsection{Updating of \texorpdfstring{$q(\bC,\btheta )$}{C,theta}}\label{eq:Section_update_autoencoder}

The relationship between $\bC,\btheta$ and $\bM$ is complex and nonlinear. As discussed before in \cref{subsec:observation_model,subsec:encoders}, we use neural networks to model $p(\bM\mid\bC,\btheta;\bw_D)$ and we aim to optimize these neural networks to minimize $\calL_1$. From \eqref{L1}, we have
\begin{dgroup*} 
	\begin{dmath*}
		\calL_1=-\E{q(\bC)q(\btheta)q(\tbC)q(\bs)}\left[{\log p(\bC \mid \tbC,\bs)}+\log {p(\bM\mid\bC,\btheta;\bw_D )}+\log p(\btheta)-\log {q(\bC)}-\log {q(\btheta)}\right]
	\end{dmath*}
	\begin{dmath}
		=-\E{q(\bC)q(\btheta)q(\tbC)q(\bs)}\left[{\log p(\bC \mid \tbC,\bs)}+\log {p(\bM\mid\bC,\btheta; \bw_D )}-\log {q(\bC)}\right]\quad\quad\\
		-\E{q(\btheta)}\left[\log p(\btheta)-\log {q(\btheta)}\right]. \label{eq:L_1}
	\end{dmath}
\end{dgroup*}
Recall that $\bw_D$, $\bw_R$ and $\bw_E$ denote the parameters of the decoder network, the reliability encoder network and the event encoder network, respectively. To learn $\bw_D$ with the gradient descent method, we need to compute the gradient of $\calL_1$ with respect to $\bw_D$. Denoting the two expectation terms of \eqref{eq:L_1} as $\calL_{11}$ and $\calL_{12}$, namely 
\begin{align}
	\calL_1=\calL_{11}+\calL_{12},\label{eq:cost_l1}
\end{align}
and we have
\begin{dmath*}
	\nabla_{\bw_D}\calL_1=\nabla_{\bw_D}\calL_{11}+\nabla_{\bw_D}\calL_{12}.
\end{dmath*}
As $\btheta$ is a discrete variable, $\calL_{12}$ is easy to compute. The gradient of $\calL_{11}$ with respect to $\bw_D$ is given by
\begin{dgroup*} 
	\begin{dmath*}
		\nabla_{\bw_D}\calL_{11} =-\nabla_{\bw_D}\E{q(\bC)q(\btheta)q(\tbC)q(\bs)}\left\{\{{\log p(\bC \mid \tbC,\bs)}+\log {p(\bM\mid\bC,\btheta;\bw_D )}-\log {q(\bC)}\}\right\}
	\end{dmath*}
	\begin{dmath*}
		=-\E{q(\bC)q(\btheta)q(\tbC)q(\bs)}\{\nabla_{\bw_D}\{{\log p(\bC \mid \tbC,\bs)}+\log {p(\bM\mid\bC,\btheta;\bw_D )}-\log {q(\bC)}\}\},\label{eq:partial_dirivative_L_11}
	\end{dmath*}
\end{dgroup*}
which can be computed using the stochastic gradient descent (SGD) algorithm. In each iteration, we replace $\nabla_{\bw_D}\calL_{11}$ with its unbiased estimator $\nabla_{\bw_D}\calL_{11}'$, where 
\begin{align}
\calL_{11}'\triangleq{\log p(\bC \mid \tbC,\bs)}+\log {p(\bM\mid\bC,\btheta;\bw_D )}-\log {q(\bC)}\label{eq: cost_autoencoder}
\end{align} 
with $\bC$, $\btheta$, $\bs$, and $\tbC$ being sampled from $q(\bC)$, $q(\btheta)$, $q(\bs)$, and $q(\tbC)$ respectively. %Thus, we have converted the optimization problem of minimizing $\calL_1$ with respect to $\bw_D$ into minimizing $\calL'_{11}+\calL_{12}$ with respect to $\bw_D$.

We cannot use the same process to deal with $\bw_R$ and $\bw_E$. This is because
\begin{dmath*}
	\nabla_{\bw_R}\calL_{11}=-\nabla_{\bw_E}\E{q(\bC)q(\btheta)q(\tbC)q(\bs)}\left[{\log p(\bC \mid \tbC,\bs)}+\log {p(\bM\mid\bC,\btheta; \bw_D )}-\log {q(\bC)}\right]
	\neq-\E{q(\bC)q(\btheta)q(\tbC)q(\bs)}\left[\nabla_{\bw_R}\{{\log p(\bC \mid \tbC,\bs)}+\log {p(\bM\mid\bC,\btheta;\bw_D )}-\log {q(\bC)}\}\right],
\end{dmath*}
as $\bw_R$ is the parameter of $q(\bC)$. The same reason applies for $\bw_E$. To obtain unbiased estimators for these variational parameters, we need to use the reparameterization trick \cite{Kin2017}. 

From \eqref{eq:bC_given_b}, we have $\vect(\bC_n)(i) \sim \N{\bo_n(i)}{b}$, where $\bo_n$ is generated by the reliability encoder network and $b$ is a hyperparameter. We can reparameterize $\bC_n(i)$ as 
\begin{align}
\zeta_n&\sim \N{0}{1},\nonumber\\
\vect(\bC_n)(i)&=\bo_n(i)+\zeta_n b.\label{eq:sample_bC'}
\end{align} 
From \eqref{eq:hbtheta_j}, $\theta_j\sim\Cat{\bu_j}$,  where the weight vector $\bu_j\triangleq (\bu_j(r))_r$ is generated by the event encoder network. Then, according to (1) in \cite{JanGuPoo2016}, we can reparameterize $\theta_j$ as 
\begin{align}
\bchi_j(r)&\sim\text{Gumbel}(0,1)\nonumber\\
\theta_j&={\argmax_{r}}[\bchi_j(r)+\log \bu_j(r)],\label{eq:sample_hbtheta}
\end{align}
where the Gumbel(0, 1) distribution can be sampled by first drawing $ \Upsilon\sim
\text{Uniform}(0, 1)$ and then computing $\bchi_j(r) = -\log(-\log(\Upsilon))$.
With $\bC_n$ and $\theta_j$ being \eqref{eq:sample_bC'} and \eqref{eq:sample_hbtheta} respectively, and letting $\bzeta=(\zeta_n)$, $\bchi=(\bchi_j)$ and $\bu=(\bu_j)$, we then obtain 
\begin{dmath}
	\nabla_{\bw_R}\calL_{11}=-\nabla_{\bw_R}\E{q(\bC)q(\btheta)q(\tbC)q(\bs)}\left[{\log p(\bC \mid \tbC, \bs)}+\log {p(\bM\mid\bC,\btheta;\bw_D )}-\log {q(\bC)}\right]\nonumber\\
	= -\E{p(\bzeta)p(\bchi)q(\tbC)q(\bs)}\left[\nabla_{\bw_R}\left\{
	\sum_{n=1}^N \log \calN(\bo_n+\zeta_n b;\\\vect(\tbC_{s_n}),b'\bI)+\log {p(\bM\mid\bzeta,\bchi;\bw_D,\bo,\bu )}\right\}\right],\label{eq: gradients_w2}
\end{dmath}
where $\calN(\cdot; \bmu,\bsigma)$ is the Gaussian probability density function with mean $\bmu$ and variance $\bsigma$. Here, $\bo=(\bo_n)$ is a function of $\bw_R$ and $\bu$ is a function of $\bw_E$. We also have 
\begin{dmath}
	\nabla_{\bw_E}\calL_{11}=-\E{p(\bzeta)p(\bchi)q(\tbC)q(\bs)}\left[\nabla_{\bw_E}\{{\log p(\bC \mid \tbC, \bs)}+\log {p(\bM\mid\bC,\btheta;\bw_D )}-\log {q(\bC)}\}\right]
	= -\E{p(\bzeta)p(\bchi)}\left[\nabla_{\bw_E}\left\{
	\log {p(\bM\mid\bzeta,\bchi;\bw_D,\bo,\bu )}\right\}\right].\label{eq: gradients_w3}
\end{dmath}
Now in each iteration, we can apply SGD to find an unbiased estimator of the gradient of  $\calL_1$ with respect to $\bw_E$ and $\bw_R$.

\begin{Remark}\label{remark1}
	The max function in equation \eqref{eq:sample_hbtheta} is not differentiable and following \cite{JanGuPoo2016}, we use the softmax function as an approximation to $\argmax$, i.e., we let 
	\begin{align}
	\mathrm{prob}(\theta_j)(r)=\dfrac{\exp(\bchi_j(r)+\log \bu_j(r)/\tau)}{\sum_{r'}\exp(\bchi_j(r')+\log \bu_j(r')/\tau)}, \label{eq:theta_tao}
	\end{align}
	where $\tau$ is the temperature parameter. 
\end{Remark}
\begin{Remark}\label{remark2}
	The term $\calL_{12}$ in \cref{eq:cost_l1} is used to constrain the distance between the variational distribution $q(\btheta)$ and the prior distribution $p(\btheta)$ of event states. When the dataset includes enough observations or useful social network information, then $\calL_{12}$ is more important in the initial iterations and the importance decreases as the number of iterations increases. In these cases, we revise \cref{eq:cost_l1} to
	\begin{align}
	\calL_1=\calL_{11}+\kappa^{i}\calL_{12},\label{eq:cost_l1_kappa}
	\end{align} 
	where $0<\kappa<1$ is a hyperparameter and $i$ denotes the $i$-th iteration.

\end{Remark}
\subsection{Updating of \texorpdfstring{$q(\tbC,\bs,\bpi,\bz,\bbeta )$}{pi,beta}}
We want to find variational parameters corresponding to $(\tbC,\bs,\bpi,\bz,\bbeta)$ to minimize $\calL_2$. To achieve this, we iteratively update these variational parameters in ART. The variational parameters corresponding to $\bbeta$ and $\bz$ are updated in the same way as our previous work \cite{YanWanTay2019}. For completeness, we reproduce the results below. 

\subsubsection{Social network parameter \texorpdfstring{$\bbeta$}{beta}}
Let $\blambda_k=(G_{k}, H_{k})$. We choose the variational distribution of $\beta_k$ to be in the same exponential family as its posterior distribution, namely $q(\beta_{k})=\Beta{G_k}{H_k}$. Similar to (14) and (15) in our previous work\cite{YanWanTay2019}, we can show that
\begin{align}
G_k&= \sum_{(n,m)}\bA(n,m)\phi_{n \rightarrow m,k}\phi_{m \rightarrow n,k}+g_0 \label{eq:update_Gk},\\
H_k&= \sum_{(n,m)}(1-\bA(n,m))\phi_{n \rightarrow m,k}\phi_{m \rightarrow n,k}+h_0 \label{eq:update_Hk},
\end{align}
where $\phi_{n \rightarrow m,k}= q(z_{n \rightarrow m}=k)$ is defined in \cref{subsec:Community membershipz}.

From (10) in \cite{Min2003}, we also have
\begin{align}
\E{q(\beta_{k})}[\log(\beta_{k})]&=\Psi(G_k)-\Psi(G_k+H_k)\text{, and } \label{eq:Eq_logBeta} \\
\E{q(\beta_{k})}[\log(1-\beta_{k})]&=\Psi(H_k)-\Psi(G_k+H_k),\label{eq:Eq_log1minBeta}
\end{align}
which are used in computing the variational distributions of other parameters in our model. Here, $\Psi(\cdot)$ is the digamma function.

\subsubsection{Community membership indicators \texorpdfstring{$\bz$}{z}}\label{subsec:Community membershipz}
We let the variational distribution of $z_{n\rightarrow m}$ to be in the same exponential family as its posterior distribution, namely a categorical distribution with probabilities $(\phi_{n\rightarrow m,k})_{k=1}^{K}$. Similar to (19) in our previous work\cite{YanWanTay2019}, one can show that if $\bA(n,m)=0$,
\begin{dmath}
	\phi_{n \rightarrow m,k}
	\propto \exp\{\phi_{m \rightarrow n,k}\left(\E{q(\beta_{k})}[\log(\beta_{k})]-\log(\epsilon)\right)+\E{q(\bpi_{n})}[\log(\pi_{n,k})]\}\label{eq:update_phi_D1},
\end{dmath}
where $\E{q(\beta_{k})}[\log(\beta_{k})]$ and $\E{q(\bpi_{n})}[\log(\pi_{n,k})]$ are computed using \eqref{eq:Eq_logBeta} and \eqref{eq:Eq_log_pi} in the sequel, respectively. On the other hand, if $\bA(n,m)=0$, we have
\begin{dmath}
	\phi_{n \rightarrow m,k}
	\propto \exp\{\phi_{m \rightarrow n,k}\left(\E{q(\beta_{k})}[\log(1-\beta_{k})]-\log(1-\epsilon)\right)+\E{q(\bpi_{n})}[\log(\pi_{n,k})]\}\label{eq:update_phi_D0},
\end{dmath}
where $\E{q(\beta_{k})}[\log(1-\beta_{k})]$ is computed in \eqref{eq:Eq_log1minBeta} in the sequel.

\subsubsection{Event community indices \texorpdfstring{$\bs$}{s}}\label{subsec:s}
We take $q(s_n)=\Cat{(\psi_{n,k})_{k=1}^{K}}$, where $(\psi_{n,k})_{k=1}^{K}$ is the variational parameter. Let $\hpsi_{n,k} \triangleq \log(\psi_{n,k})$. We have 
\begin{align*}
&p(s_n=k \mid \bC_n,\tbC_k,\bpi_n)\\
&\propto p(s_n=k\mid\bpi_n)p(\bC_n \mid \tbC_k)\\
&=\pi_{n,k}\prod_{r_1=1}^{R_1}\prod_{r_2=1}^{R_2}\N{\bC_n(r_1,r_2);  \tbC_k(r_1,r_2)}{b'}.
\end{align*}
Thus, $p(s_n=k \mid \bC_n,\tbC_k,\bpi)$ is a categorical distribution and is in the same exponential family as $q(s_n)=\Cat{(\psi_{n,k})_{k=1}^{K}}$.
Let 
\begin{align*}
\varsigma_{n,k}&\triangleq \log \left(\pi_{n,k}\prod_{r_1=1}^{R_1}\prod_{r_2=1}^{R_2}\N{\bC_n(r_1,r_2);  \tbC_k(r_1,r_2)}{b'}\right).\\
\varsigma_{n,1:K}&\triangleq\log\left(\sum_{k=1}^{K}\exp(\varsigma_{n,k})\right).
\end{align*}
The natural parameter of $p(s_n \mid \bC_n,\tbC_k,\bpi)$ is 
\begin{align}
(\varsigma_{n,k}-\varsigma_{n,1:K})_{k=1}^K. \label{term:natural_parameter_psn}
\end{align}
Assume the natural parameter of $q(s_n)$ is
\begin{align} 
(\hpsi_{n,k}-\hpsi_{n,1:K})_{k=1}^K,\label{term:natural_parameter_qsn}
\end{align}
where $\hpsi_{n,1:K}\triangleq\log\left(\sum_{k=1}^{K}\exp(\hpsi_{n,k})\right)$.
Acording to the relationship between the natural gradient and  and the natural parameters (i.e., (22) in \cite{HofBleWan2013}), the natural gradient of $\calL_3$ in \eqref{eq:L3_s} with respect to $\hpsi_{n,k}$ can be derived from \eqref{term:natural_parameter_psn} and \eqref{term:natural_parameter_qsn} and the result is
\begin{dmath*}
	\hnabla_{\hpsi_{n,k}}\calL_3
	=\hpsi_{n,k}-\hpsi_{n,1:K}-\E{q\left(\bC_n,\tbC_k,\bpi_{n} \right)}\left[\varsigma_{n,k}-\varsigma_{n,1:K}\right].
\end{dmath*}

We sample $\bC_n$ from $q(\bC_n)$ and obtain the unbiased estimator of $\hnabla_{\hpsi_{n,k}}\calL_3$ as
\begin{align}
&\hnabla_{\hpsi_{n,k}}\calL'_3= \hpsi_{n,k}\nonumber\\
&-\sum_{r_1=1}^{R_1}\sum_{r_2=1}^{R_2}\E{q\left(\tbC_k(r_1,r_2) \right)}\left[\log\N{\bC_n(r_1,r_2);\tbC_k(r_1,r_2)}{b'}\right]\nonumber\\
&-\E{q(\bpi_n)}\left[\log\pi_{n,k}\right]-\triangle_n, \label{eq: L_3'}
\end{align}
where $\triangle_n\triangleq \hpsi_{n,1:K}-\E_{q\left(\bC_n,\tbC_k,\bpi_{n} \right)}[\varsigma_{n,1:K}]$ and it is constant for $k=1,\cdots,K$.
Then we update $\hpsi_{n,k}$ using 
\begin{align}
\hpsi_{n,k}^{(i)}=\hpsi_{n,k}^{(i-1)}-\rho^{(i)}\hnabla_{\hpsi_{n,k}}\calL'_3,\label{eq:update_tpsi}
\end{align}
where $\rho^{(i)}$ is the known step size at $i$-th iteration. Let $\Xi_{n,k}$ be the first three terms of the right-hand side of \eqref{eq: L_3'}. We compute exponential function of both sides of \eqref{eq:update_tpsi} and obtain
\begin{dmath}
	\psi_{n,k}^{(i)}=\psi_{n,k}^{(i-1)} \exp\left(-\rho^{(i)} \Xi_{n,k}^{(i)}\right) \exp(\rho^{(i)}\triangle_n^{(i)})\\
	=\psi_{n,k}^{(i-1)}\exp\left\{-\rho^{(i)}\hpsi_{n,k}^{(i-1)}-\rho^{(i)}\sum_{r_1=1}^{R_1}\sum_{r_2=1}^{R_2}\dfrac{1}{2{b'}^2}\left((\tbsigma_k^{(i-1)}(r_1,r_2))^2+\left(\tbmu_k^{(i-1)}(r_1,r_2)-\bC_n^{(i)}(r_1,r_2)\right)^2\right)+\rho^{(i)}\E{q(\bpi_n)}\left[\log\pi_{n,k}\right]\right\} \exp(\rho^{(i)}\triangle_n^{(i)})	
\end{dmath}
where $\tbmu$ and $\tbsigma$ are variational parameters of $\tbC$ defined in Section \ref{sec:tbC} and $\E{q(\bpi_{n})}\left[\log\pi_{n,k}\right]$ can be computed by \eqref{eq:Eq_log_pi} below.
Since $\sum_{k=1}^{K}\psi_{n,k}^{(i)}=1$ and $\exp(\rho^{(i)}\triangle_n^{(i)})$ is constant for every $\psi_{n,k}^{i}$,  $k=1,\cdots,K$, in each iteration, we only need to compute $\psi_{n,k}^{(i-1)} \exp\left(-\rho^{(i)} \Xi_{n,k}^{(i)}\right)$ and then 
\begin{align}
\psi^{(i)}_{n,k}=\dfrac{\psi_{n,k}^{(i-1)} \exp\left(-\rho^{(i)} \Xi_{n,k}^{(i)}\right)}{\sum_{k=1}^{K}\psi_{n,k}^{(i-1)} \exp\left(-\rho^{(i)} \Xi_{n,k}^{(i)}\right)}\label{eq:update_psi}. 
\end{align}

\subsubsection{Community reliability matrix \texorpdfstring{$\tbC$}{C}}\label{sec:tbC}
We let the variational distribution of $\tbC_k$ for each $k=1,\ldots,K$ to be given by
\begin{align}\label{eq:q_tbC}
q(\tbC_k(r_1,r_2))=\N{\tbmu_k(r_1,r_2)}{\tbsigma_k(r_1,r_2)}. 
\end{align}
We also have
\begin{dmath}
	p(\tbC_k(r_1,r_2)\mid \bs, \{\bC_n(r_1,r_2)\}_n)\propto p(\tbC_k(r_1,r_2))\prod_{\{n:s_n=k\}}{p(\bC_n(r_1,r_2) \mid \tbC_k(r_1,r_2))},
\end{dmath}
which is a normal distribution with mean (cf.\ \cref{eq:bC_b',eq:prior_tbC_k})
\begin{dmath*}
	\dfrac{1}{\bV(r_1,r_2)^{-2}+\frac{1}{b'}\sum_n I(s_n,k)}\left(\dfrac{\bU_k(r_1,r_2)}{\bV_k(r_1,r_2)^2}+\frac{1}{b'}\sum_n\bC_n(r_1,r_2)I(s_n,k)\right)
\end{dmath*}
and variance 
\begin{align*}
\dfrac{1}{\bV_k(r_1,r_2)^{-2}+\frac{1}{b'}\sum_n I(s_n,k)}.
\end{align*}
Let $\hbmu_k(r_1,r_2)\triangleq\dfrac{\tbmu_k(r_1,r_2)}{\tbsigma_k(r_1,r_2)^2}$ and $\hbsigma_k(r_1,r_2)^2\triangleq-\dfrac{1}{2\tbsigma_k(r_1,r_2)^2}$. As $q(\tbC_k(r_1,r_2))$ is in the same exponential family as $p(\tbC_k(r_1,r_2)\mid \bs, \{\bC_n(r_1,r_2)\}_n)$ and its natural parameter is $(\hbmu_k(r_1,r_2),\hbsigma_k(r_1,r_2))$. Similar to \cref{subsec:s}, we sample $\bC_n$ from $q(\bC_n)$ and update $\hbmu_k(r_1,r_2)$ and $\hbsigma_k^2(r_1,r_2)$ using
\begin{dgroup}
	\begin{dmath}
		\hbmu_k^{(i)}(r_1,r_2)=\hbmu_k^{(i-1)}(r_1,r_2)-\rho^{(i)}\left(\hbmu_k^{(i-1)}(r_1,r_2)\\ {-\dfrac{\bU_k(r_1,r_2)}{\bV_k(r_1,r_2)^2} -\frac{1}{b'}\sum_n\bC_n^{(i)}(r_1,r_2)\psi^{(i)}_{n,k}}\right),\quad\quad\label{eq:update_tmu}
	\end{dmath}
	\begin{dmath}
		\hbsigma_k^{(i)}(r_1,r_2)^2=\hbsigma_k^{(i-1)}(r_1,r_2)^2-\rho^{(i)}\left(\hbsigma_k^{(i-1)}(r_1,r_2)^2\quad\quad\\+\dfrac{1}{2}\left(\bV_k(r_1,r_2)^{-2}+\frac{1}{b'}\sum_n \psi^{(i)}_{n,k}\right)\right).\quad\quad\label{eq:update_tsigma}
	\end{dmath}
\end{dgroup}
Finally, we obtain
\begin{align*}
\tbsigma_k^{(i)}(r_1,r_2)^2 &=-\frac{1}{2\hbsigma_k^{(i)}(r_1,r_2)^2},\\
\tbmu_k^{(i)}(r_1,r_2)&\triangleq{\hbmu_k^{(i)}(r_1,r_2)}{\tbsigma_k^{(i)}(r_1,r_2)^2}.
\end{align*} 

\subsubsection{Mixture weights \texorpdfstring{$\bpi$}{pi}}
We let $q(\bpi_n)=\Dir{\gamma_n}$ and thus $q(\bpi_n)$ is an exponential family distribution and its variational parameter $\gamma_n$ is also its natural parameter. To find the variational parameter $\bgamma$  that minimizes $\calL_2$, we find the partial derivative 
\begin{dmath}
	\nabla_{\bgamma}\calL_2=-\nabla_{\bgamma}\left\{\E{q(\bpi)q(\bs)q(\bz)}\left[{\log p(\bs \mid \bpi)}+{\log p(\bz\mid\bpi)}+\log p(\bpi)-\log q(\bpi)\right]\right\}
	=-\nabla_{\bgamma}\left\{\E{q(\bpi)q(\bs)q(\bz)}\left[{\log p(\bpi\mid \bs, \bz)} -\log q(\bpi)\right]\right\},\label{eq:frac_L_gamma}
\end{dmath}
where 
\begin{align*}
&p\left(\bpi_n\mid \{s_i\}_{i=1}^N, \{z_{n\rightarrow m}\}_{m=1,m\neq n}^N \right)\\
%\propto& p(\bpi_n,s_n^l,\bz)\\
&\propto \prod_{i=1}^N p(s_i\mid \bpi_n)\prod_{m=1,m\neq n}^{N}p(z_{n\rightarrow m}\mid \bpi_n)p(\bpi_n)\\
&\propto \Dir{\left(\frac{\alpha}{K}+\sum_{m=1,m\neq n}^{N}I(z_{n\rightarrow m},k)+\sum_{i=1}^{N}I(s_i,k)\right)_{k=1}^{K}}.
\end{align*}
Recall that $K$ represents the maximum number of communities. As $q(\bpi_k)=\Dir{\gamma_k}$ is in the same exponential family as $p\left(\bpi_n\mid s_n, \{z_{n\rightarrow m}\}_{m=1,m\neq n}^N \right)$, thus if we let \eqref{eq:frac_L_gamma} be zero, we obtain
\begin{dmath}
	\gamma_{n,k}=\E{q(s_n,\{z_{n\rightarrow m}\}_{m=1,m\neq n}^N)}\left[\frac{\alpha}{K}+\sum_{m=1,m\neq n}^{N}I(z_{n\rightarrow m},k)+\sum_{i=1}^{N}I(s_i,k)\right]
	=\frac{\alpha}{K}+\sum_{m=1,m\neq n}^{N}\phi_{n\rightarrow m,k}+ \sum_{i=1}^{N}\psi_{i,k} \label{eq:update_gamma}.
\end{dmath}
From (10) in \cite{Min2003}, we also have 
\begin{align}
\E{q(\bpi_{n})}[\log(\pi_{n,k})]=\Psi(\gamma_{n,k})-\Psi\left(\sum_{k=1}^{K}\gamma_{n,k}\right),\label{eq:Eq_log_pi}
\end{align}
which is used in \eqref{eq:update_psi}. Recall that $\Psi(\cdot)$ is the digamma function.

\section{Experimental Results}\label{sec:ExperimentResults}

In this section, experiments on three real datasets are presented\footnote{Code: https://github.com/yitianhoulai/ART}. We adopt majority voting, BCC \cite{KimGha2012}, CBCC \cite{VenGuiKaz2014}, VISIT \cite{YanWanTay2019}, TruthFinder\cite{YinHanPhi2008}, AccuSim\cite{DonBerSri2009}, GTM\cite{ZhaHan2012}, CRH\cite{LiLiGao2014}, CATD\cite{ LiLiGao2015a}, and KDEm \cite{WanCheKap2016} as the state-of-the-art benchmark methods. We test the performance of different methods on the IMDB dataset augmented with Twitter information, which we have made available in \cite{YanTay2018}. We also test the performance of different methods on the Sentiment Polarity (SP) dataset and the Weather Sentiment (CF) dataset. The last two datasets do not provide network information. To emulate network information in our experiments, for any pair of agents, we calculate the observation differences of their commonly observed events. Next, we calculate the mean of the absolute values of the differences and assume the pair of agents are connected if the mean is less than or equal to 0.2. 

\subsection{Description of Datasets}\label{sec:Datasets}

\subsubsection{Sentiment Polarity (SP) Dataset}

The Sentiment Polarity dataset \cite{Venanzi2015} contains agent classifications for movie comments. All the movie comments are from the website Rotten Tomatoes.\footnote{\url{https://www.rottentomatoes.com/}} Agents were asked to annotate each comment as either positive (1) or negative (0). The uncertainty or confidence of the agents' annotations are however unavailable. Ground truth labels are from experts. The dataset contains 27,746 evaluations from 203 agents on 4,999 movie comments (i.e., events). 

\subsubsection{IMDB Dataset}\label{sec:IMDB_4}

We collected data from the website IMDB\footnote{\url{https://www.imdb.com/}} and Twitter \cite{YanTay2018}. If a user rates a movie in IMDB and clicks the share button, a Twitter message is generated. We collected movie evaluations from IMDB and social network information from Twitter. We divide the movie evaluations into 2 levels: bad (0-5), good(6-10). We treat the ratings on the IMDB website, which are based on the aggregated evaluations from all users, as the event truths whereas our observations come from only a subset of users who share their ratings on Twitter. To better show the influence of social network information on event truth discovery, we delete small subnetworks that have less than 5 agents each. The final dataset \cite{YanTay2018} we use consists of 2266 evaluations from 209 individuals on 245 movies (events) and also the social network between these 209 individuals. Similar to \cite{For2010,GopBle2013}, we regard the social network to be undirected as both follower or following relationships indicate that the two users have similar taste.

\subsubsection{Weather Sentiment (CF) Dataset}\label{sec:IMDB_2}
The Weather Sentiment dataset was provided by Crowd-Flower (CF) and can be found in \cite{Venanzi2015}. The agents were asked to classify the tweets with respect to the weather sentiment into the following categories: negative (0), neutral (1), positive (2), tweet not related to weather (3) and cannot tell (4). The dataset contains 1,720 evaluations from 461 agents on 300 tweets (i.e., events).

\subsection{Experiment Settings}\label{sec:Experiment_Settings}
%<*tag:Experiment-settings> 
The hyperparameters in our proposed ART method are tuned using another dataset called MS from \cite{Venanzi2015}. Since the hyperparameters of our method are tuned with this dataset, we do not compare the performance of ART and the benchmark methods on this dataset. The values of the hyperparameters in our method are given in \cref{table:experiment_settings}. We use the same set of hyperparameters except $\kappa$ to do experiments on all the three datasets in \cref{sec:Datasets}.  
%</tag:Experiment-settings>	

\begin{table}[!htb]
	\caption{Experiment Settings}
	\centering
	\begin{tabular}{L{6cm}|L{2cm}}
		\hline
		Hyper-parameter & Value \\ \hline
		$\epsilon$ in \eqref{eq:prior_Dnm}     & $10^{-10}$   \\ \hline
		$b$ in \eqref{eq:bC_given_b} and $\bb'$ in \eqref{eq:bC_b'}    & 0.1	\\ \hline
		Each element of $ \bU_k$ in \eqref{eq:prior_tbC_k}     & Random value between 0.4 and 0.5  \\ \hline
		Each element of $\bV_k$ in \eqref{eq:prior_tbC_k}     & $0.1$  \\ \hline
		$\tau$ in \eqref{eq:theta_tao} & 0.01 \\ \hline
		Maximum number of communities $K$ & 3 \\ \hline
		Size of the reliability matrices $R_1\times R_2$ & 6$\times$3 \\ \hline
		Sizes of 2 hidden layers of the reliability encoder &  128, and 32 \\ \hline
		Sizes of 2 hidden layers of the event encoder  & 128, and 32\\ \hline
		Sizes of 3 hidden layers of the decoder& 16, 32, and 128  \\ \hline
		Learning rate of ADAMS optimizer & 0.001 \\ \hline		
		Step size at $i$-th iteration & $0.1$\\ \hline
		$\kappa$ in \cref{eq:cost_l1_kappa} & IMDB and SP: 0.9; CF:1.0\\ \hline
	\end{tabular}
	\label{table:experiment_settings}
\end{table}

\subsection{Sensitivity Analysis}\label{subsec:Sensitivity_analysis}
We conduct experiments on the three datasets using the same set of hyperparameters except $\kappa$ in \cref{eq:cost_l1_kappa}. The CF dataset contains only 1,720 observations and has no social network information and thus the majority voting prior is important in all the learning iterations. We thus set $\kappa$ be 1.0.  For the IMDB and SP datasets, we set $\kappa$ to be 0.9 since the  SP dataset is rich in observations and the IMDB dataset contains social network information. For these two datasets, the importance of the majority voting prior decreases as the iteration progresses. The results in \cref{subsec:Metric_and_result} suggest that our method is not sensitive to the hyperparameters and can achieve the comparable accuracies on different datasets using the same set of hyperparameters except $\kappa$. In the following, we perform experiments to show the influence of different hyperparameters on our method.

\subsubsection{Impact of reliability matrice size \texorpdfstring{$R_1\times R_2$}{R1xR2}}\label{sec:ART_different_C}
ART uses reliability matrices to represent the reliabilities of agents. In this experiment, we study impact of 4 different reliability matrix sizes on the truth discovery accuracy.
%<*performance-rm-size>
The results show that the accuracy of ART is related to the number of elements (i.e., the product of $R_1$ and $R_2$) in the reliability matrices. We observe that reliability matrices of size $2\times 9$ and $3\times 6$ achieve the same accuracies on all three real datasets. However, we also observe that different sizes of the matrix all achieve comparable results.
%</performance-rm-size> 

\begin{table}[!htb]
	\caption{Truth discovery accuracy of ART with different reliability matrix sizes.} % title of Table
	\centering % used for centering table
	\begin{tabular}{l|lll}
		\hline
		$R_1\times R_2$	& SP                & IMDB             & CF             \\ \hline
		2$\times$ 9        & \textbf{0.918} & \textbf{0.771} & \textbf{0.903}          \\ \hline
		6$\times$ 3       & \textbf{0.918}  & \textbf{0.771} & \textbf{0.903}          \\ \hline
		3$\times$ 3        & 0.914           & 0.767           & 0.890          \\ \hline
		6$\times$ 6        & 0.915           & 0.763          & 0.897         \\ \hline		
	\end{tabular}
	\label{table:acc_RealData_size} % is used to refer this table in the text
\end{table}

\subsubsection{Impact of number of communities \texorpdfstring{$K$}{K}}\label{sec:ART_different_k}
In this experiment, we show the impact of the number of communities $K$ on the truth discovery accuracy of ART. From \cref{table:acc_RealData_k}, we observe that ART achieves relatively good accuracies on all three datasets with different number of communities.
\begin{table}[!htb]
	\caption{Truth discovery accuracy of ART with different number $K$ of communities.} % title of Table
	\centering % used for centering table
	\begin{tabular}{l|lll}
		\hline
		$k$	& SP                & IMDB             & CF             \\ \hline
		3        & \textbf{0.918} & 0.771 & \textbf{0.903}          \\ \hline
		6        & 0.915  & 0.759 & 0.9          \\ \hline
		9        & 0.915           & \textbf{0.788}           &  0.89         \\ \hline			
	\end{tabular}
	\label{table:acc_RealData_k} % is used to refer this table in the text
\end{table}

\subsubsection{Impact of variance \texorpdfstring{$\bV_k$}{Vk}}\label{sec:ART_different_V}
In ART, we use $\bV_k$ to denote the variance of the prior distribution of $\tbC_k$. From \cref{table:acc_RealData_V}, we observe that different $\bV_k$ have larger influence on CF than the other two datasets.  ART achieves the best accuracies on all the three datasets when each element of $\bV_k$ is 0.1.
\begin{table}[!htb]
	\caption{Truth discovery accuracy with different $\bV_k$.} % title of Table
	\centering % used for centering table
	\begin{tabular}{l|lll}
		\hline
		Each element of $\bV_k$	& SP                & IMDB             & CF             \\ \hline
		0.01        & 0.916 			& \textbf{0.771} & 0.867          \\ \hline
		0.1        	& \textbf{0.918}  & \textbf{0.771} & \textbf{0.903}         \\ \hline
		1        	& 0.917           & \textbf{0.771}           &  0.877         \\ \hline			
	\end{tabular}
	\label{table:acc_RealData_V} % is used to refer this table in the text
\end{table}

\subsection{Comparison with Benchmarks}\label{subsec:Metric_and_result}
We evaluate and compare the performance of ART and the benchmark methods using five metrics: accuracy, area under the ROC curve (AUC), precision, recall, and the F1 score. The top performer and those close to it (i.e, within 0.002 of the top) are highlighted in boldface in the subsequent tables. The following results indicate that ART is competitive, and often superior, compared to the other benchmark methods. However, the computation time of ART is larger than most of the other benchmark methods (except VISIT) since we include the agent network information in our inference. Our deep autoencoder is also more computationally expensive to learn.

\subsubsection{Accuracy}
The accuracies of ART and the benchmark methods are shown in \cref{table:Acc_RealData1}. It can be observed that ART achieves the best accuracies on all the three datasets.
\begin{table}[!htb]
	\caption{Truth discovery accuracy.} % title of Table
	\centering % used for centering table
	\begin{tabular}{l|lll}
		\hline
		Method      & SP             & IMDB           & CF             \\ \hline
		ART         & \textbf{0.918} & \textbf{0.771} & \textbf{0.903} \\ \hline
		VISIT       & \textbf{0.916} & 0.751          & 0.893          \\ \hline
		CBCC        & \textbf{0.916} & 0.714          & 0.893          \\ \hline
		BCC         & 0.915          & 0.678          & 0.890          \\ \hline
		MV          & 0.885          & 0.710          & 0.867          \\ \hline
		TruthFinder & 0.885          & 0.714          & 0.880          \\ \hline
		AccuSim     & 0.890          & 0.702          & 0.730          \\ \hline
		GTM         & 0.890          & 0.710          & 0.743          \\ \hline
		CRH         & 0.894          & 0.706          & 0.747          \\ \hline
		CATD        & 0.873          & 0.657          & 0.807          \\ \hline
		KDEm        & 0.890          & 0.702          & 0.877          \\ \hline
	\end{tabular}
	\label{table:Acc_RealData1} % is used to refer this table in the text
\end{table}

\subsubsection{Area under the ROC Curve (AUC)}
The AUC scores of ART and the benchmark methods are shown in \cref{table:Auc_RealData1}. For the multi-class case (i.e., the CF dataset), AUC scores are calculated for each label, and their average is taken. It can be observed that ART achieves the best AUC score on the SP dataset and is among one of the best methods on the CF dataset. On the IMDB dataset, VISIT and BCC achieve the best AUC scores. 
\begin{table}[!htb]
	\caption{Truth discovery AUC.} % title of Table
	\centering % used for centering table
	\begin{tabular}{l|lll}
		\hline
		Method      & SP             & IMDB           & CF             \\ \hline
		ART         & \textbf{0.958} & 0.844          & \textbf{0.948} \\ \hline
		VISIT       & \textbf{0.958} & \textbf{0.871} & 0.913          \\ \hline
		CBCC        & \textbf{0.957} & 0.859          & \textbf{0.949} \\ \hline
		BCC         & \textbf{0.957} & \textbf{0.872} & \textbf{0.950}  \\ \hline
		MV          & 0.885          & 0.759          & 0.865          \\ \hline
		TruthFinder & 0.885          & 0.765          & 0.873          \\ \hline
		AccuSim     & 0.890           & 0.753          & 0.754          \\ \hline
		GTM         & 0.890           & 0.759          & 0.763          \\ \hline
		CRH         & 0.894          & 0.756          & 0.764          \\ \hline
		CATD        & 0.873          & 0.708          & 0.824          \\ \hline
		KDEm        & 0.890           & 0.753          & 0.861          \\ \hline
	\end{tabular}
	\label{table:Auc_RealData1} % is used to refer this table in the text
\end{table}

\subsubsection{Precision}
The precision scores of ART and the benchmark methods are shown in \cref{table:precision_RealData1}. For the multi-class case (i.e., the CF dataset), precisions are calculated for each label, and their average, weighted by the number of true instances for each label, is taken.  It can be observed that ART achieves the best precision on the IMDB dataset and is one of the best methods on the SP dataset. On the CF dataset, CBCC achieves the best precision. 
\begin{table}[!htb]
	\caption{Truth discovery precision.} % title of Table
	\centering % used for centering table
	\begin{tabular}{l|lll}
		\hline
		Method      & SP             & IMDB           & CF             \\ \hline
		ART         & \textbf{0.921} & \textbf{0.639} & 0.878          \\ \hline
		VISIT       & \textbf{0.922} & 0.607          & 0.896          \\ \hline
		CBCC        & \textbf{0.922} & 0.568          & \textbf{0.902} \\ \hline
		BCC         & \textbf{0.92}  & 0.535          & 0.865          \\ \hline
		MV          & 0.868          & 0.563          & 0.873          \\ \hline
		TruthFinder & 0.868          & 0.566          & 0.885          \\ \hline
		AccuSim     & 0.886          & 0.556          & 0.748          \\ \hline
		GTM         & 0.874          & 0.563          & 0.762          \\ \hline
		CRH         & 0.884          & 0.559          & 0.758          \\ \hline
		CATD        & 0.863          & 0.519          & 0.824          \\ \hline
		KDEm        & 0.878          & 0.556          & 0.879          \\ \hline
	\end{tabular}
	\label{table:precision_RealData1} % is used to refer this table in the text
\end{table}

\subsubsection{Recall}
The recall scores of ART and the benchmark methods are shown in \cref{table:recall_RealData1}. For the multi-class case (i.e., the CF dataset), recall scores are calculated for each label, and then their average, weighted by the number of true instances for each label, is taken. It can be observed that ART achieves the best recall on two datasets: the SP and CF datasets. On the IMDB dataset, TruthFinder achieves the best recall. 
\begin{table}[!htb]
	\caption{Truth discovery recall.} % title of Table
	\centering % used for centering table
	\begin{tabular}{l|lll}
		\hline
		Method      & SP             & IMDB           & CF             \\ \hline
		ART         & \textbf{0.913} & 0.867          & \textbf{0.903} \\ \hline
		VISIT       & 0.910          & 0.911          & 0.893          \\ \hline
		CBCC        & 0.910          & 0.933          & 0.893          \\ \hline
		BCC         & 0.909          & 0.944          & 0.890           \\ \hline
		MV          & 0.907          & 0.944          & 0.867          \\ \hline
		TruthFinder & 0.907          & \textbf{0.956} & 0.880           \\ \hline
		AccuSim     & 0.894          & 0.944          & 0.730          \\ \hline
		GTM         & \textbf{0.911} & 0.944          & 0.743          \\ \hline
		CRH         & 0.907          & 0.944          & 0.747          \\ \hline
		CATD        & 0.887          & 0.900          & 0.807          \\ \hline
		KDEm        & 0.906          & 0.944          & 0.877          \\ \hline
	\end{tabular}
	\label{table:recall_RealData1} % is used to refer this table in the text
\end{table}

\subsubsection{F1 score }
The F1 scores of ART and the benchmark methods are shown in \cref{table:f1_RealData1}. For the multi-class case (i.e., the CF dataset), F1 scores are calculated for each label, and their average, weighted by the number of true instances for each label, is taken. This averaging takes label imbalance into consideration and can result in an F1 score that is not between the corresponding precision and recall. It can be observed that ART achieves the best F1 scores on two datasets: the IMDB and SP datasets. On the CF dataset, ART is slightly worse than VISIT, which achieves the best F1 score. 
\begin{table}[!htb]
	\caption{Truth discovery F1 score.} % title of Table
	\centering % used for centering table
		\begin{tabular}{l|lll}
		\hline
		Method      & SP             & IMDB           & CF             \\ \hline
		ART         & \textbf{0.917} & \textbf{0.736} & 0.890          \\ \hline
		VISIT       & \textbf{0.916} & 0.729          & \textbf{0.894} \\ \hline
		CBCC        & \textbf{0.916} & 0.706          & 0.884          \\ \hline
		BCC         & 0.914          & 0.683          & 0.877          \\ \hline
		MV          & 0.887          & 0.705          & 0.868          \\ \hline
		TruthFinder & 0.887          & 0.711          & 0.882          \\ \hline
		AccuSim     & 0.890          & 0.700          & 0.712          \\ \hline
		GTM         & 0.892          & 0.705          & 0.731          \\ \hline
		CRH         & 0.895          & 0.702          & 0.730          \\ \hline
		CATD        & 0.875          & 0.659          & 0.814          \\ \hline
		KDEm        & 0.891          & 0.700          & 0.876          \\ \hline
	\end{tabular}
	\label{table:f1_RealData1} % is used to refer this table in the text
\end{table}

\subsubsection{Average execution time}\label{sec:execution_time}
We tabulate the average execution time of ART and the benchmark methods in \cref{table:execution_time}. All the methods are run on the same laptop with a core-i5-8250U CPU and 16G RAM. 

The execution time of ART is over 100 times larger than most of the benchmark methods since ART includes the agent network information and the deep autoencoder is computationally expensive to learn. The SP dataset contains the largest number of evaluations and the CF dataset contains the largest number of agents and thus experiments on these two datasets take more time. Another benchmark that uses the agent network information is VISIT, which however has longer average execution times than ART due to the complex variational inference approach used in its non-conjugate observation model.  ART is therefore more suitable for offline truth discovery applications, where inference performance is more important than latency.
\begin{table}[!htb]
	\caption{Average execution time.} % title of Table
	\centering % used for centering table
	\begin{tabular}{@{}l|lll@{}}
		\toprule
		Method      & SP           & IMDB         & CF           \\ \midrule
		ART         & 1261.1       & 92.0         & 118.8        \\
		VISIT       & 1413.6       & 108.5         & 562.1       \\
		CBCC        & 4.0          & 2.4          & 4.2          \\
		BCC         & 5.1          & 1.7          & 1.9          \\
		MV          & 1.0          & 0.1          & 0.1          \\
		TruthFinder & \textbf{0.4} & \textbf{0.1} & \textbf{0.1} \\
		AccuSim     & 41.9         & 3.2          & 0.8          \\
		GTM         & 1.0          & 0.1          & 0.1          \\
		CRH         & 1.5          & 0.1          & 0.1          \\
		CATD        & 0.5          & 0.1          & 0.1          \\
		KDEm        & 0.9          & 0.1          & 0.1          \\ \bottomrule
	\end{tabular}
	\label{table:execution_time} % is used to refer this table in the text
\end{table}

\section{Conclusion}\label{sec:conclusion}

In this paper, we have combined the strength of autoencoders in learning nonlinear relationships and the strength of Bayesian networks in characterizing hidden interpretable structures to tackle the truth discovery problem. The Bayesian network model introduces constraints to the autoencoder and at the same time incorporates the community information of the social network into the autoencoder. We  developed a variational inference method to estimate the parameters in the autoencoder and infer the hidden variables in the Bayesian network. Results on real datasets demonstrate the competitiveness of our proposed method over other state-of-the-art benchmark methods. 

In this paper, we have not considered correlations between events in our inference. We have also not incorporated any side information or prior information about the events into our procedure. These are interesting future research directions, which may further improve the truth discovery accuracy.

%\red{[Many conference names are not consistent or not capitalized properly.]}

\bibliographystyle{IEEEtran}
\bibliography{IEEEabrv,DeepTruthDiscovery}

\end{document}

%% file: preamble.tex
\usepackage[T1]{fontenc}
\usepackage{amsmath,amssymb,amsfonts,mathrsfs,bm}% Typical maths resource packages
\usepackage{mathtools}
\usepackage{amsthm}
\usepackage[shortlabels]{enumitem}
\usepackage{graphicx}
\usepackage{epstopdf}
\usepackage{url}
\usepackage{colortbl}
\usepackage{multirow}
\usepackage{xcolor}
\usepackage[normalem]{ulem}
\usepackage{array}
\newcolumntype{L}[1]{>{\raggedright\let\newline\\\arraybackslash\hspace{0pt}}m{#1}}
\newcolumntype{C}[1]{>{\centering\let\newline\\\arraybackslash\hspace{0pt}}m{#1}}
\newcolumntype{R}[1]{>{\raggedleft\let\newline\\\arraybackslash\hspace{0pt}}m{#1}}

\makeatletter
\let\MYcaption\@makecaption
\makeatother
\usepackage[font=footnotesize]{subcaption}
\makeatletter
\let\@makecaption\MYcaption
\makeatother

\usepackage{xparse}



\usepackage{glossaries}
\newacronym{wrt}{w.r.t.}{with respect to}
\newacronym{RHS}{R.H.S.}{right-hand side}
\newacronym{LHS}{L.H.S.}{left-hand side}
\newacronym{iid}{i.i.d.}{independent and identically distributed}
\usepackage{float}

\ifx\notloadhyperref\undefined
	\ifx\loadbibentry\undefined
		\usepackage[hidelinks,hypertexnames=false]{hyperref} 
	\else
		\usepackage{bibentry}
		\makeatletter\let\saved@bibitem\@bibitem\makeatother
		\usepackage[hidelinks,hypertexnames=false]{hyperref}
		\makeatletter\let\@bibitem\saved@bibitem\makeatother
	\fi
\else
	\relax
\fi

\usepackage[capitalize]{cleveref}
\crefname{equation}{}{}
\Crefname{equation}{}{}
\crefname{claim}{claim}{claims}
\crefname{step}{step}{steps}
\crefname{line}{line}{lines}
\crefname{dmath}{}{}
\crefname{dseries}{}{}
\crefname{dgroup}{}{}
\crefname{Theorem}{Theorem}{Theorems}
\crefname{Corollary}{Corollary}{Corollaries}
\crefname{Proposition}{Proposition}{Propositions}
\crefname{Lemma}{Lemma}{Lemmas}
\crefname{Definition}{Definition}{Definitions}
\crefname{Example}{Example}{Examples}
\crefname{Assumption}{Assumption}{Assumptions}
\crefname{Remark}{Remark}{Remarks}
\crefname{Rem}{Remark}{Remarks}
\crefname{remarks}{Remarks}{Remarks}
\crefname{Theorem_A}{Theorem}{Theorems}
\crefname{Corollary_A}{Corollary}{Corollaries}
\crefname{Proposition_A}{Proposition}{Propositions}
\crefname{Lemma_A}{Lemma}{Lemmas}
\crefname{Definition_A}{Definition}{Definitions}

\usepackage{algorithm,algorithmic}

\ifx\loadbreqn\undefined
	\relax
\else
	\usepackage{breqn} 
\fi

\ifx\renewtheorem\undefined
\ifx\useTheoremCounter\undefined
\newtheorem{Theorem}{Theorem}
\newtheorem{Corollary}{Corollary}
\newtheorem{Proposition}{Proposition}

\else

\fi

\newtheorem{Remark}{Remark}

\fi

\theoremstyle{remark}

\theoremstyle{plain}

\newcommand{\calB}{\mathcal{B}}

\newcommand{\calL}{\mathcal{L}}

\newcommand{\calN}{\mathcal{N}}

\newcommand{\calS}{\mathcal{S}}

\newcommand{\bA}{\mathbf{A}}
\newcommand{\bb}{\mathbf{b}}

\newcommand{\bC}{\mathbf{C}}
\newcommand{\bd}{\mathbf{d}}

\newcommand{\bI}{\mathbf{I}}

\newcommand{\bM}{\mathbf{M}}

\newcommand{\bo}{\mathbf{o}}

\newcommand{\bp}{\mathbf{p}}

\newcommand{\bs}{\mathbf{s}}

\newcommand{\bu}{\mathbf{u}}
\newcommand{\bU}{\mathbf{U}}

\newcommand{\bV}{\mathbf{V}}
\newcommand{\bw}{\mathbf{w}}

\newcommand{\bx}{\mathbf{x}}

\newcommand{\bz}{\mathbf{z}}

\DeclareSymbolFont{bsfletters}{OT1}{cmss}{bx}{n}
\DeclareSymbolFont{ssfletters}{OT1}{cmss}{m}{n}
\DeclareMathSymbol{\bsfGamma}{0}{bsfletters}{'000}
\DeclareMathSymbol{\ssfGamma}{0}{ssfletters}{'000}
\DeclareMathSymbol{\bsfDelta}{0}{bsfletters}{'001}
\DeclareMathSymbol{\ssfDelta}{0}{ssfletters}{'001}
\DeclareMathSymbol{\bsfTheta}{0}{bsfletters}{'002}
\DeclareMathSymbol{\ssfTheta}{0}{ssfletters}{'002}
\DeclareMathSymbol{\bsfLambda}{0}{bsfletters}{'003}
\DeclareMathSymbol{\ssfLambda}{0}{ssfletters}{'003}
\DeclareMathSymbol{\bsfXi}{0}{bsfletters}{'004}
\DeclareMathSymbol{\ssfXi}{0}{ssfletters}{'004}
\DeclareMathSymbol{\bsfPi}{0}{bsfletters}{'005}
\DeclareMathSymbol{\ssfPi}{0}{ssfletters}{'005}
\DeclareMathSymbol{\bsfSigma}{0}{bsfletters}{'006}
\DeclareMathSymbol{\ssfSigma}{0}{ssfletters}{'006}
\DeclareMathSymbol{\bsfUpsilon}{0}{bsfletters}{'007}
\DeclareMathSymbol{\ssfUpsilon}{0}{ssfletters}{'007}
\DeclareMathSymbol{\bsfPhi}{0}{bsfletters}{'010}
\DeclareMathSymbol{\ssfPhi}{0}{ssfletters}{'010}
\DeclareMathSymbol{\bsfPsi}{0}{bsfletters}{'011}
\DeclareMathSymbol{\ssfPsi}{0}{ssfletters}{'011}
\DeclareMathSymbol{\bsfOmega}{0}{bsfletters}{'012}
\DeclareMathSymbol{\ssfOmega}{0}{ssfletters}{'012}

\newcommand{\balpha}{\bm{\alpha}}
\newcommand{\bbeta}{\bm{\beta}}
\newcommand{\bgamma}{\bm{\gamma}}

\newcommand{\btheta}{\bm{\theta}}
\newcommand{\bmu}{\bm{\mu}}

\newcommand{\bpi}{\bm{\pi}}

\newcommand{\bsigma}{\bm{\sigma}}

\newcommand{\bzeta}{\bm{\zeta}}

\newcommand{\bchi}{\bm{\chi}}
\newcommand{\bphi}{\bm{\phi}}
\newcommand{\bpsi}{\bm{\psi}}

\newcommand{\blambda}{\bm{\lambda}}

\newcommand{\bLambda}{\bm{\Lambda}}

\newcommand{\bOmega}{\bm{\Omega}}

\newcommand{\hpsi}{\widehat{\psi}}

\DeclareMathOperator*{\argmax}{arg\,max}

\DeclareMathOperator{\vect}{vec}

\DeclarePairedDelimiter\braces{\{}{\}}

\newcommand{\qednew}{\nobreak \ifvmode \relax \else
      \ifdim\lastskip<1.5em \hskip-\lastskip
      \hskip1.5em plus0em minus0.5em \fi \nobreak
      \vrule height0.75em width0.5em depth0.25em\fi}

\newcommand{\bone}{\mathbf{1}}

\newcommand{\cond}[2]{\left. {#1}\, \middle| \, {#2} \right.}

\DeclareDocumentCommand \P { g d() g } {%
	\IfNoValueTF {#3} 
	{%
		\IfNoValueTF {#1} 
		{%
			\IfNoValueTF {#2}
			{%
				\mathbb{P}%
			}%
			{%
				\mathbb{P}\left({#2}\right)%
			}%
		}%
		{%
			\IfNoValueTF {#2}
			{%
				\mathbb{P}_{#1}%
			}%
			{%
				\mathbb{P}_{#1}\left({#2}\right)%
			}%		
		}%
	}%
	{%
		\IfNoValueTF {#1} 
		{%
			\mathbb{P}\left(\cond{#2}{#3}\right)%
		}%
		{%
			\mathbb{P}_{#1}\left(\cond{#2}{#3}\right)%
		}%	
	}%
}

\DeclareDocumentCommand \E { g o g } {%
	\IfNoValueTF {#3} 
	{%
		\IfNoValueTF {#1} 
		{%
			\IfNoValueTF {#2}
			{%
				\mathbb{E}%
			}%
			{%
				\mathbb{E}\left[{#2}\right]%
			}%
		}%
		{%
			\IfNoValueTF {#2}
			{%
				\mathbb{E}_{#1}%
			}%
			{%
				\mathbb{E}_{#1}\left[{#2}\right]%
			}%		
		}%
	}%
	{%
		\IfNoValueTF {#1} 
		{%
			\mathbb{E}\left[\cond{#2}{#3}\right]%
		}%
		{%
			\mathbb{E}_{#1}\left[\cond{#2}{#3}\right]%
		}%	
	}%
}

\newcommand{\Bern}[1]{\mathrm{Bern}\left(#1\right)}

\newcommand{\Dir}[1]{\mathrm{Dir}\left(#1\right)}
\newcommand{\Cat}[1]{\mathrm{Cat}\left(#1\right)}
\newcommand{\N}[2]{{\calN\left({#1},\: {#2}\right)}}
\newcommand{\Beta}[2]{{\calB e\left({#1},\: {#2}\right)}}

\definecolor{gray90}{gray}{0.9}

\ifx\nohighlights\undefined

	\newcommand{\msout}[1]{\text{\color{green} \sout{\ensuremath{#1}}}}
	\newcommand{\del}[1]{{\color{green}\ifmmode \msout{#1}\else\sout{#1}\fi}}
\else

	\newcommand{\msout}[1]{#1}
	\newcommand{\del}[1]{#1}
\fi

\newcommand{\hide}[1]{}
\graphicspath{{./Figures/}} 
\ifx\diagnoselabel\undefined
	\relax
\else
	\makeatletter
	 \def\@testdef #1#2#3{%
		 \def\reserved@a{#3}\expandafter \ifx \csname #1@#2\endcsname
		\reserved@a  \else
	 \typeout{^^Jlabel #2 changed:^^J%
	 \meaning\reserved@a^^J%
	 \expandafter\meaning\csname #1@#2\endcsname^^J}%
	 \@tempswatrue \fi}
	\makeatother
\fi